# MatchMiner-AI: Open-source, Privacy-preserving Cancer Clinical Trial Matching using Artificial Intelligence

Jennifer Altreuter, MD; Pavel Trukhanov, MSc; Morgan A. Paul, MB; Michael J. Hassett, MD, MPH; Irbaz B. Riaz, MD PhD; Muhammad Umar Afzal, MBBS; Arshad A. Mohammed B.A.; Ayub Umair, PhD; Huan He, PhD; Chueh Husan Hsu, MD; Sarah Sammons, MD; James Lindsay, PhD; Emily Mallaber, BA; Harry R. Klein, PhD; Gufran Gungor, BS; Matthew Galvin, BS; Michael D'Eletto, MS; Sabrina Y. Camp, MS; Stephen C. Van Nostrand, MS; James Provencher, MS; Joyce Yu, MBA; Naeem Tahir, MD; Alexandra S. Bailey, MD; Jonathan Wischhusen, MD; Olga Kozyreva, MD; Taylor Ortiz, MD; Hande Tuncer, MD; Jad El Masri, MD; Alys Malcolm, MD; Tali Mazor, PhD; Ethan Cerami, PhD; Kenneth L. Kehl, MD, MPH

Affiliations: Dana-Farber Cancer Institute (JA, PT, MAP, MJH, SS, JL, EM, HK, GG, MG, MD, SYC, SCVN, JP, JY, NT, ASB, JW, OK, TO, HT, JEM, AM, TM, EC, KLK); Mayo Clinic (IR, MUA, AAM, AU, HH, CHH)

**Abstract**

Background: Clinical trials are essential to advancing cancer treatments, but fewer than 10% of adults with cancer enroll in therapeutic trials. Open-source AI trial matching tools could democratize access to trial options. Methods: We created MatchMiner-AI, co-developed with practicing clinical oncologists and trained on synthetic electronic health record (EHR) data. It uses open-weight LLMs to summarize patient histories from unstructured EHR text and extract target populations from trial eligibility documents. Embedding and re-ranking models were distilled to retrieve and rank trial and patient suggestions. Multifaceted evaluation was performed, including retrospective quantification of distillation fidelity; applying a closed-source LLM as judge of patient summarization and matching; and evaluation of candidate matches by oncologists. Results: Across retrospective evaluations of distillation fidelity, the pipeline outperformed a baseline text-embedding model, improving mean average precision (MAP) at 20 from 0.44 (95% CI 0.44-0.45) to 0.95 (95% CI 0.95-0.96) for trial-enrolled patients and from 0.38 (95% CI 0.37-0.38) to 0.94 (95% CI 0.93-0.94) for patients who received standard of care therapies. In a 50-patient sample selected for comparison between MatchMiner-AI and a rules-based tumor genomic trial matching algorithm, MatchMiner-AI retrieved trials for all patients, as opposed to 19 patients (38%) who had tumor genomic data available. Among those 19 patients, 80% of 256 trial suggestions retrieved by MatchMiner-AI were deemed reasonable considerations by a frontier LLM, vs 53% of 113 suggestions retrieved by the rules-based approach. Conclusion: MatchMiner-AI is an open-source, open-weights, clinical trial matching AI pipeline for oncology. Synthetic training data, model weights, inference tools, and demonstration frontends are publicly available.

**Funding**

Meta Corp Llama Impact Grant; Nancy Lurie Marks Family Foundation; United States NIH/NCI (R37CA295653); United States Department of Defense (W81XWH2210086).

## Introduction

Clinical trials are critical for developing new cancer treatments and improving patient outcomes. Historically, however, under 10% of adults with cancer in the United States have participated in treatment trials.[1] At the same time, many trials fail to reach their accrual goals.[2,3] Oncology clinical trials are complex, often involving multiple arms and specific treatment and biomarker requirements.[4,5] The time and resources required for an individual oncologist to navigate complex eligibility criteria present a major barrier to finding an appropriate trial for a patient. A need has therefore arisen for methods that can match patients to clinical trials at scale.

Several such tools, initially leveraging structured clinical data, have been developed in recent years.[6–14] Most recently, efforts have been undertaken to apply large language models (LLMs) to unstructured clinical data for clinical trial matching. This has included using LLMs to extract structured data from clinical patient summaries[15–18] and/or clinical trial criteria.[15–17,19] Other groups have published more comprehensive LLM-based trial matching systems.[20,21] Prominent examples included TrialGPT, which used keyword matching for trial retrieval and a commercial LLM to evaluate eligibility and rank trials given a single patient document;[17] and PRISM/Triomics, which is a commercial system that structures trial eligibility evaluation as a series of hierarchical clinical questions, retrieving relevant chunks of text from the patient's longitudinal record to answer each using a proprietary LLM.[22]

In general, many trial matching tools attempt to extract all eligibility criteria for a trial and evaluate if a patient matches each one. This comprehensive approach is intuitive, but given the number and complexity of eligibility criteria for modern cancer trials, criterion-by-criterion AI pre-screening may risk constraining the utility of automated trial-matching pipelines. For tasks like feasibility analysis or filtering through thousands of trial options, rapid retrieval based on core criteria, such as age, sex, cancer type, disease context, treatment history, and key biomarkers, may be at least as valuable.

Indeed, platforms utilizing AI for healthcare must consider real world constraints to technology adoption, including integration into existing practice workflows and electronic health record (EHR) systems, as well as ease of product implementation and maintenance. Pipelines that assume a patient's relevant history has been summarized into a single document or a set of pre-specified structured fields may have limited utility when applied to raw longitudinal EHR data. In a busy oncology practice, it is also essential to minimize the time from patient lookup to the delivery of potential matches. Trial matching platforms which require the implementation of new data models, data dictionaries or additional software products may be difficult to maintain. A simple, shareable, well-engineered solution is necessary to move from proof of concept to practice.

Relying on commercial models and/or closed-source tools for trial matching therefore has drawbacks. These may include incentives to prioritize matches to trials run by for-profit funders and data distribution drift resulting from rapid evolution in frontier models. Commercial models have also become increasingly expensive to access, and they can be deprecated or withdrawn anytime. They also require private patient information to be sent to a third-party vendor over the Internet.[23] Open-weights models and open-source pipelines can therefore provide a cost-effective alternative.

For developers who wish to create open-source clinical trial matching tools, however, data security and patient privacy constraints create critical barriers. When models are trained on protected health information (PHI), they can "memorize" and later regenerate it or expose it to membership inference attacks.[24] Federated learning, in which training data remain at source institutions, but model weights and updates are still shared, does not inherently solve this problem. Models and pipelines developed using fully synthetic data, without grounding in PHI,[25,26] could be an alternative.

In this study, we sought to develop a novel, open-source, AI-based platform for cancer clinical trial matching based on fully synthetic EHR data, overcoming restrictions on sharing solutions grounded in individual-level PHI, and to assess the performance of this platform using real-world data among patients receiving cancer treatment. The platform was developed to integrate with EHR data, directly works with EHR text -- with no requirement for external data models, dictionaries, or ontologies -- and can provide real-time matches for oncologists within their typical daily workflows.

## Methods

### MatchMiner-AI overview

The overall MatchMiner-AI approach **(Figure 1)** included modules to (1) extract the core eligibility criteria (defined as age, sex, cancer type/histology, cancer burden/treatment intent, biomarker requirements, and prior treatment requirements) for target patient populations, or "spaces," for each trial; (2) summarize longitudinal patient histories from unstructured EHR text using an LLM; (3) rank candidate patient-trial "space" matches; (4) predict, on a 0–5 ordinal scale, whether and how specifically a given candidate patient meets the core eligibility criteria for a specific trial; and (5) predict whether a patient meets common exclusion criteria, such as uncontrolled brain metastases or a history of a recent second primary cancer, for a specific trial unrelated to the core eligibility criteria (a "boilerplate exclusions" check).

### MatchMiner-AI training components

*Clinical trial "spaces"*

The clinicaltrials.gov Application Programming Interface (API) was queried in October 2025 to identify active, recruiting, or not yet open trials for malignant condition terms (including cancer, carcinoma, sarcoma, lymphoma, leukemia, myeloma, and myelodysplastic, myeloproliferative) listed as Phase I, II, III, or IV studies. For each trial, the Gemma-4-31B-IT LLM was prompted to extract a list of clinical "spaces" for the trial from its eligibility criteria, where each space was defined as a unique combination of core clinical concepts (age, sex, cancer type, histology, burden of disease, prior treatment, and biomarkers) that might render the patient eligible (prompt in **Supplemental Table 2**). Some trials have only one "space," or target population, whereas others, such as basket or umbrella trials, have several. The output was parsed to extract the text describing each target clinical space. The prompt also instructed the LLM to extract "boilerplate" exclusion criteria at the trial level, defined as a history of pneumonitis, heart failure, renal dysfunction, liver dysfunction, uncontrolled brain metastases, HIV or hepatitis, or poor performance status. Each trial could therefore have more than one "space" but only one set of boilerplate exclusion criteria.

Three clinical trial datasets were generated. These included one training set and two evaluation sets: 1) a random sample of 3,000 trials that had no enrollments at our center from January 2016 (corresponding to initial current EHR implementation at our center) through April 2024 (data lock date), used for training; 2) a non-overlapping random sample of 500 trials that did not enroll at our center, used for distillation fidelity evaluation among patients at our center who started standard of care treatments during that window; and 3) all trials with patient enrollments at our center during that window, used for distillation fidelity evaluation among patients who enrolled on any trial during that time.

*Synthetic EHR data generation*

Within the 3,000-trial training sample, for each clinical trial “space” as described above, the Gemma-4-31B teacher LLM was prompted to create semi-structured text histories for 5 hypothetical patients who met the space-defining eligibility criteria and 5 patients who “did not quite” meet the criteria. The prompt directed the LLM to create text histories in the form of lists of 20-30 plausible clinical events along the disease trajectory. Event types included new diagnoses, systemic therapy starts, surgeries, radiation treatments, adverse events, clinical progress notes, imaging reports, pathology reports, and next-generation sequencing (NGS) reports. Next, for each event in these text histories corresponding to a clinical document (progress notes, imaging reports, pathology reports, and NGS reports), the Gemma-4-31B teacher LLM was prompted to invent a plausible full-length document based on the event, in the context of the proceeding and subsequent events for that synthetic patient. To promote more realistic note generation, text augmentation steps were applied in which generic anti-cancer drug names were randomly replaced with their brand names or with common abbreviations; and in which, for 10% of patients, a prior history of a second primary cancer was incorporated. Prompts are provided in **Supplemental Table 2.** The synthetic data pipeline is depicted in **Supplemental Figure 1.**

*Summarizing patient histories*

We deployed the Gemma-4-31B-IT LLM to summarize each patient’s full longitudinal history, based on their free text document histories. To accommodate records that exceed the model’s context window, each patient’s record was processed in chronological, overlapping chunks (approximately 50,000 tokens with a 500-token overlap); the model maintained a structured running summary that was updated as each successive chunk was processed, yielding the final patient summary after the last chunk. The patient summarization prompt (**Supplemental Table 2**) instructed the LLM to generate semi-structured output capturing the same core clinical concepts used to define trial spaces. The prompt also included instructions to summarize any history of conditions that might meet common clinical trial "boilerplate" exclusion criteria, including uncontrolled brain metastases, lack of measurable disease, congestive heart failure, pneumonitis, renal dysfunction, liver dysfunction, and HIV or hepatitis infection. For any such boilerplate exclusion condition, the LLM was prompted to provide evidence of the condition from the patient's history.

*LLM-based patient-trial candidate evaluation*

During training, retrospective candidate patient-space combinations were defined by linking each free-text synthetic patient summary based on a target population from a given real clinical trial to each of the free-text clinical spaces extracted for that trial. We then prompted the Gemma-4-31B-IT teacher LLM to evaluate whether each trial space was a "reasonable consideration" based on the patient summary. The prompt (**Supplemental Table 2**) instructed the LLM to first reason about whether the patient matched the core clinical criteria, and then to generate a 0–5 ordinal trial match quality score: 1 point for the trial being a reasonable consideration (defined as the patient not being clearly excluded based on the core clinical criteria requirements), plus 1 additional point each for the trial specifying the patient's cancer type, disease burden/stage, prior treatment, and biomarker status (full rubric and prompt in **Supplemental Table 2**). A score of 0 indicated the trial was not a reasonable consideration. This match quality standard did not include the "boilerplate" exclusion criteria, which were handled by a separate prompt to Gemma-4-31B-IT.

*"TrialSpace" text embedding model development*

The Qwen3-0.6B-Embedding[27] text embedding model was fine-tuned to embed patient summaries and trial “spaces” into the same mathematical vector space, enabling subsequent rapid retrieval. The text embedding model was fine-tuned simultaneously on two objectives: (1) using the multiple negatives ranking loss[28] to discriminate between true patient-space combinations that passed the minimum “reasonable consideration” standard per the Gemma-4-31B prompt above and random patient-space combinations; and (2) using the CoSENT loss[29] to place patient-space combinations with higher Gemma-generated match quality scores (normalized from their original 0-5 range to the -1 to 1 cosine similarity range) closer together in embedding space than those with lower scores.

We empirically found that the initial TrialSpace training step yielded a model that could identify clinical trials based on cancer type but did not discriminate well according to specific treatment history or biomarker criteria. Therefore, we fine-tuned TrialSpace further to improve performance at discriminating among trial spaces or patients that were initially highly ranked matches but were not “reasonable considerations.” We used the preliminary TrialSpace model to identify the 20 trial spaces that best matched each patient summary and the 40 patient summaries that best matched each trial space, based on the cosine similarity metric. Gemma-4-31B was again prompted to score each of these top patient/trial-space match candidates. This intermediate dataset was therefore enriched for candidate combinations that were more reasonable considerations. We then further fine-tuned TrialSpace a second time, on the same tasks, using this enriched dataset. The process was then repeated a third and final time to improve performance at discriminating among otherwise highly ranked candidate matches.

*“TrialChecker” re-ranking/regression model development*

The TrialSpace model enables pre-calculation of embedding vectors for patient summaries and trial spaces, enabling subsequent queries of each to be performed rapidly without requiring inference on each combination of patient and trial space.[30] However, single document-level vectors can sacrifice granularity, and cosine similarity between patient and trial space vectors is not a clinically intuitive metric for oncologists and trial investigators. To improve specificity and interpretability, we therefore distilled a "TrialChecker" text re-ranking/regression model by fine-tuning ModernBERT-large[31] to predict the Gemma-4-31B-IT 0–5 ordinal trial match quality score for a candidate patient-trial space combination (trained as a regression objective, with the TrialChecker 0-1 output range re-scaled to the 0–5 range). TrialChecker is therefore a cross-encoder model[32] that can be applied just to the top ranked TrialSpace matches. TrialChecker was applied to filter out matches predicted not to be reasonable and prioritize higher-scoring candidates for display to the user.

*"BoilerplateChecker" classification model development*

For all unique patient-clinical trial combinations ranked during each iteration of TrialSpace training, elements in the patient summary describing conditions that often exclude patients from trials in general (e.g., uncontrolled brain metastases or a history of pneumonitis) were extracted and linked to the corresponding "boilerplate" exclusion criteria for each trial. The Gemma-4-31B-IT teacher LLM was then prompted to determine whether any of the patient's history matched any of these boilerplate trial exclusion criteria. If so, a positive outcome label (indicating ineligibility/exclusion) was assigned. Using these labels, a "BoilerplateChecker" text classification model was distilled onto ModernBERT-large to generate a predicted probability of exclusion based on a combination of patient and trial boilerplate text strings.

*Clinical piloting and iterative development*

An initial version of the pipeline was deployed for four medical oncologists on our study team. Based on their feedback during monthly meetings over a six-month pilot period, the pipeline was modified and re-trained. Examples of this feedback and changes implemented in response are provided in **Supplemental Table 1.** The updated pipeline was then piloted with eight total medical oncologists, including the four initial users. All eight oncologists have a general scope of practice at our community-based regional sites.

## Evaluation

### *Distillation fidelity*

<u>Dana-Farber Cancer Institute (DFCI) patient cohorts</u>

EHR data were obtained using the DFCI Oncology Data Retrieval System[34] for all adults who (a) enrolled in cancer treatment clinical trials and had consent dates recorded in the institutional OnCore database from January 2016 to April 2024, and/or who (b) initiated standard of care systemic therapy regimens during that time. These data included unstructured clinical notes, imaging reports, and pathology reports. Access to data for this study was approved by the Dana/Farber Harvard Cancer Center (DF/HCC) Institutional Review Board (IRB). Given the large volume of data required, minimal risk to patients, and infeasibility of re-contacting patients in this retrospective study, waivers of informed consent were obtained. For the trial-enrolled denominator, trials open at the time of the patient's enrollment were eligibile for retrieval. For the standard of care denominator, the random sample of 500 trials that were not open at DFCI during that window was evaluated. For both denominators, therefore, trials eligible for retrieval were not included in training.

Evaluation was performed using the full retrospective trial enrollee dataset and a 10% patient-level random sample of the standard of care dataset, given the much larger size of the latter cohort. In the trial enrollment dataset, the task was not to evaluate eligibility for the specific trial on which the patient enrolled, but rather to measure performance of distilled models at predicting Gemma-4-31B's responses on the trial match quality and boilerplate exclusion tasks, given a set of trial options, using the patient history through the time of an actual trial enrollment event.

Distillation fidelity metrics

The full pipeline was then evaluated on real clinical data for two use cases: (1) Finding relevant trials for individual patients, and (2) finding relevant patients for individual trials. The primary distillation fidelity metrics were precision at N=20 trial spaces (use case 1) and at N=40 patients (use case 2), defined as the proportion of the top 20 or top 40 pipeline-surfaced match candidates respectively that were at least a reasonable consideration (score at least 1 on the ordinal 0-5 scale) per Gemma-4-31B-IT; and mean average precision (MAP) at N=20 and N=40 respectively. MAP was calculated by taking the mean across observations of the average precision (AP) @ N=20 and N=40 respectively, where AP @ N represents, for each observation, the mean of the observation's precision @ k metrics, calculated for each k from 1 to N.[35] These metrics were calculated first using all match candidates retrieved by the TrialSpace embedding model, and then again after filtering for match candidates that were predicted to be at least a reasonable consideration (score at least 0.2 on the normalized 0-1 scale, corresponding to at least 1 on the teacher LLM 0-5 scale) by TrialChecker. To evaluate TrialSpace's performance against a general domain text embedding model, the metrics were also calculated using Llama-Embed-Nemotron-8B[36] (which has ten times the parameter count of TrialSpace) for embedding and retrieval.

Among TrialSpace-retrieved match candidates, TrialChecker distillation fidelity was further evaluated both by calculating binary classification metrics (area under receiver operating characteristic curve, AUROC; area under precision-recall curve, AUPRC; and calibration curves) for a binary outcome of whether the candidate was ‘reasonable’ (score >= 1); and by calculating regression metrics, including R-squared (coefficient of determination), root mean squared error (RMSE), mean absolute error (MAE), and Pearson’s R-statistic, against the ordinal 0-5 LLM outcome. BoilerplateChecker distillation fidelity for its binary outcome was also evaluated using the area under the receiver operating characteristic curve (AUROC), area under the precision-recall curve (AUPRC), and calibration curves.

Bootstrap 95% confidence intervals were generated for all distillation fidelity statistics. Statistics were also calculated by demographic strata, including patient age, sex, race, and ethnicity.

*LLM-as-judge and medical oncologist reviews*

A randomly selected sample of 50 patients with progressive disease (as ascertained by separate text classification models running on imaging reports and medical oncologist notes)[39] at our center was identified in May 2026. For this sample, Claude Opus 4.8 and Gemini 3.5 Flash were prompted to judge the quality of the patient summaries generated by Gemma-4-31B-IT, with options including ‘AGREE’ (no clinically material differences); OMISSION (Gemma summary omits information relevant to trial eligibility); INCORRECT (Gemma summary included information not backed by the EHR documents); or OMISSION_AND_INCORRECT (prompt in Supplemental Table 2). Similarly, Opus and Gemini was prompted to judge the quality of a patient’s trial match candidates and boilerplate exclusion checks as generated using MatchMiner-AI, using the same scoring rubric given to Gemma-4-31B-IT originally; agreement with Gemma's scores was quantified using rates of exact and (for trial match quality's ordinal scale) within-1 agreement.

For a random sample of 20 patients in this denominator, a sample of 5 trial spaces from the top 20 TrialSpace-retrieved target populations was randomly selected for manual annotation of match quality by the study team. Five oncologists participated, each annotating 25 patient–trial space pairs; the trial spaces for 5 of the 20 patients (25 patient–trial space pairs) were assigned to two oncologists to enable inter-rater reliability calculations. For each candidate match, oncologists recorded Yes/No answers to five clinical alignment questions: (1) "Is this trial at least a reasonable consideration for this patient — i.e., the patient does not clearly meet an exclusion criterion such as wrong cancer type, wrong age group, wrong sex, or an excluded biomarker?; (2) Does the trial specify the patient's cancer type/histology (e.g., 'breast cancer', 'non-small cell lung cancer') rather than being open to any/all cancers (e.g., 'solid tumors', 'advanced cancers'), and does it match this patient?; (3) Does the trial have specific prior-treatment requirements (e.g., 'must have progressed on platinum-based chemotherapy', 'prior immunotherapy required') that this patient's treatment history satisfies? (Ignore washout-period timing.); (4) Does the trial require a specific biomarker (e.g., 'EGFR mutation', 'PD-L1 ≥ 50%', 'HER2-positive') that this patient is known to have?; (5) Does the trial specify a particular disease stage/burden (e.g., 'metastatic', 'locally advanced', 'relapsed/refractory', a risk index) that matches this patient's disease status? (Excludes ECOG performance status and measurable-disease criteria.)" Oncologist inter-rater reliability on answers to these questions was calculated using percent agreement and the Cohen's kappa statistic, as was inter-rater reliability among Opus 4.8, Gemma-4-31B-IT, Gemma-4-E2B-IT, and oncologist annotation.

*Computational requirements*

TrialSpace, TrialChecker, and BoilerplateChecker have under one billion parameters each and can be run on CPUs or consumer GPUs. Gemma-4-31B-IT in quantized forms can be run on a high-end consumer GPU (eg NVIDIA 5090). In production, trial-space embeddings are pre-computed, so retrieval for a new patient reduces to a nearest-neighbor lookup returning in seconds. Generating a patient summary from raw notes takes on the order of three minutes; once summaries are pre-computed, ranked trial candidates return in seconds. In deployment, patient summaries are pre-computed, and a lightweight web interface is surfaced directly via a link in the production EHR, so treating oncologists retrieve trial options on demand without additional infrastructure.

*Data and code availability*

Synthetic data, model weights, and demonstration apps are available at http://huggingface.co/ksg-dfci . Code for training and inference is available at https://github.com/dfci/matchminer-ai-training and https://github.com/dfci/matchminer-ai-inference respectively. Our retrospective DFCI evaluation datasets included protected health information (PHI) and cannot be shared. The distilled task-specific models are released in the ksg-dfci “0526” model collection on Hugging Face (TrialSpace-0526, TrialChecker-0526, and BoilerplateChecker-0526).

*Role of the funding source*

The funders of the study had no role in study design, data collection, data analysis, data interpretation, or writing of the report.

## Results

*Pipeline development and iteration*

As described above, we piloted the initial pipeline with eight oncologists practicing at DFCI-operated community-based sites. These oncologists were invited to provide qualitative feedback on the tool as accessed through a web-based frontend. Examples of this feedback and refinements implemented in response are listed in **Supplemental Table 1.** The following results describe our pipeline after these updates.

*Clinical trial "spaces" and synthetic training data*

We identified 13,160 active phase I-IV cancer clinical trials on clinicaltrials.gov, which yielded 34,519 extracted trial spaces. A subset of 3,000 trials (7,695 spaces) was randomly sampled for pipeline training to align with available computational resources. For each space, Gemma-4-31B-IT was prompted to invent longitudinal sequences of clinical events for 5 patients who met core eligibility criteria without being excluded by its 'boilerplate' exclusion criteria and 5 patients who almost, but "did not quite," meet those criteria (prompts in **Supplemental Table 2).** After excluding outputs with non-compliant formatting, this yielded 76,935 synthetic patient histories. For each clinical event in those histories corresponding to an imaging report, pathology report, oncologist assessment, or genomic sequencing report, Gemma-4-31B-IT was prompted to invent a full-length clinical document based on the longitudinal history. This yielded 1,295,607 synthetic documents.

*Retrospective distillation fidelity evaluation: DFCI patient cohorts*

For model evaluation on real patient records from our institution, we identified 7,982 therapeutic trial enrollment events for 7,056 patients and 11,393 standard-of-care treatment starts for 5,758 patients that met our eligibility criteria. Patient characteristics are listed in **Table 1.** Denominators of trials not included in training were then identified. For DFCI clinical trial participants, this included 814 trials (2,990 spaces) open at DFCI at the time of each patient's actual enrollment. Among DFCI patients who started standard of care treatments, we used a random sample of 500 trials (1,347 spaces) that were not open at DFCI at all.

Across the patient-centric and trial-centric use cases, the finetuned TrialSpace model outperformed the baseline text embedding model in retrieving relevant clinical trials for DFCI patients and identifying top candidate patients for specific trial spaces (**Figure 2**). For the trial-enrolled cohort, the ModernBERT TrialChecker reproduced the teacher's binarized "reasonable consideration" label with an area under the receiver operating characteristic curve (AUROC) of 0.95 (95% CI 0.95–0.95) in the patient-centric and 0.98 (95% CI 0.98–0.98) in the trial-centric use case; agreement with the teacher's full 0–5 ordinal score was strong (Spearman ρ 0.84, 95% CI 0.84–0.84; and 0.81, 95% CI 0.81–0.81 respectively). The BoilerplateChecker reproduced the teacher "boilerplate exclusion" judgment with AUROC 0.95 (95% CI 0.95–0.96) and 0.95 (95% CI 0.95-0.95) for the patient-centric and trial-centric use cases respectively. Metrics stratified by patient age, sex, race, and ethnicity were broadly consistent across subgroups (**Supplemental Tables 5-6).**

For the standard-of-care treatment cohort, the ModernBERT TrialChecker reproduced the teacher's binarized "reasonable consideration" label with an AUROC of 0.92 (95% CI 0.92–0.92) in the patient-centric and 0.96 (95% CI 0.96–0.96) in the trial-centric use case; agreement with the teacher's full 0–5 ordinal score remained strong (Spearman ρ 0.82, 95% CI 0.82–0.82; and 0.85, 95% CI 0.84–0.85, respectively). The BoilerplateChecker reproduced the teacher's "boilerplate exclusion" judgment with AUROC 0.97 (95% CI 0.97–0.98) in the patient-centric and 0.97 (95% CI 0.97–0.97) in the trial-centric use case. Metrics stratified by patient demographics remained consistent across subgroups (**Supplemental Tables 5–6**).

*LLM-as-judge and medical oncologist reviews*

Two commercial models (Claude Opus 4.8 and Gemini 3.5 Flash) served as a judge of Gemma-4-31B-IT's outputs on patient summarization and trial match quality scoring tasks. For patient summarization, summaries for 50 patients were scored against source EHR notes. For cases in which the commercial model judged the Gemma summary incorrect, a medical oncologist reviewed the summaries and outputs, serving as a human judge of Claude and Gemini outputs. After oncologist adjudication, 44 of 50 (88%, 95% CI 76%–94%) were judged clinically correct by Opus and Gemini each, and the remaining summaries contained a defect. Defects were concentrated in the treatment-history and biomarker sections. Examples of incorrect/omitted summary components included omission of a conditioning regimen prior to cellular therapy; misinterpretation of individual drug names in regimen acronyms (motivating prompt tweaks to add definitions of those acronyms to summaries); and failure to identify explicitly stated but incorrect documentation in source notes, including for clinically important biomarker results (eg, DpD deficiency) that had discordant reporting in the EHR. Of note, however, manual review of discrepancies also identified cases in which the commercial LLMs made the same type of error in overly trusting source documentation, but Gemma did not (**Supplemental Table 7)**.

For trial match quality scoring ("trial checking"), the Opus 4.8 judge and the production pipeline agreed exactly on the 0–5 score for 65% of 100 patient–trial pairs (95% CI 55%–74%) and within one point for 91% (95% CI 84%–95%; quadratic-weighted κ 0.74, 95% CI 0.59–0.84; Spearman ρ 0.75, 95% CI 0.60–0.87), with 95% agreement (95% CI 89%–98%; κ 0.86, 95% CI

0.71–0.97) on the binary question of whether a trial was a reasonable consideration. Gemini 3.5 Flash agreed exactly with the pipeline in 72% of pairs (95% CI 63%–80%), and within one point for 92% of pairs (95% CI 85%–96%; quadratic-weighted κ 0.76, 95% CI 0.62–0.88; Spearman ρ 0.76, 95% CI 0.61–0.88); it had 92% agreement (95% CI 85%–96%) and κ 0.79 (95% CI 0.62–0.92) for the binary reasonable consideration outcome. For boilerplate exclusion, Opus–pipeline agreement was 88% (95% CI 80%–93%; κ 0.69, 95% CI 0.51–0.84); Gemini-pipeline agreement was 82% (95% CI 73%–88%; κ 0.51, 95% CI 0.31–0.69); **Supplemental Table 7**.

A sample of 20 patient summaries was taken from the 50-patient sample above for independent dual annotation of trial match score components (overall 'reasonableness', and specific matching on cancer type, treatment history, biomarkers, burden of disease) by two oncologists and three LLMs (Claude Opus 4.8; Gemma-4-31B-IT, which as above served as the teacher model for distillation; and Gemma-4-E2B-IT, representing a small open-weights model that can be served on edge devices). For each patient summary, five of the top 20 trial spaces retrieved by TrialSpace were randomly sampled for annotation without pre-filtering by TrialChecker. Across all score components, two oncologists agreed with one another numerically more often (91%, 95% CI 86%-96%) than any judge agreed with the index oncologist (84%, 95% CI 78%-90%), and the three judges did not differ detectably from one another, spanning under two percentage points across 125 concept pairs (McNemar's exact test, all $p \geq 0.68$; **Supplemental Table 8**). All three judges were more self-consistent on test–retest than the oncologists were with one another. Scoring judges against the two-oncologist consensus increased every judge's agreement, indicating that a material share of apparent judge error was human disagreement. Notably, concept-level analysis in the full sample indicated that LLM judges often disagreed with oncologist annotators about whether a trial space specifically targeted a biomarker known to be present in the patient's tumor. This appeared to relate to differential interpretation of the question "Does the trial require a specific

biomarker…that the patient's tumor is known to have?", with LLM judges interpreting the compound question as intended, but human oncologists answering primarily the initial component of the question regarding whether the trial required a specific biomarker.

*Comparison with rules-based tumor genomic data trial matching*

The original MatchMiner tool developed at our center, here called MatchMiner-Genomics for clarity, matches patients to trials using structured tumor next-generation sequencing (NGS) results, so it can only be applied to patients who have such tumor NGS data available. In a sample of 50 patients with AI-ascertained progression of disease in May 2026, only 19 patients (38%) had such a sample. MatchMiner-AI does not require genomic data and therefore expands the denominator eligible for matching from 19 to the full 50-patient cohort. On the shared 19-patient denominator, when scored by an independent frontier-LLM oracle (Claude Opus 4.8), MatchMiner-AI surfaced at least one reasonable trial for 100% of patients (19/19), while MatchMiner-Genomics did so for 74% (14/19), and 80% (204/256) versus 53% (60/113) of trial-level suggestions were rated at least a reasonable consideration.

## Discussion

This study describes the development and evaluation of MatchMiner-AI, which is trained on synthetic EHR text to rank and evaluate cancer clinical trial options for patients and vice versa, based on core clinical criteria derived from unstructured EHR data and trial documentation. The open pipeline, built on open-weight LLMs, facilitates unbiased, reproducible trial and patient search. The synthetic data training strategy, with no seeding of data generation using individual-level PHI, allows models to be shared without risking patient privacy. The large corpus of synthetic EHR text may also be useful for other tasks, including model distillation or benchmarking. As general-domain and/or healthcare-specific open-weights LLMs improve, our synthetic data can also be re-generated, and pipeline re-trained, to leverage these improvements at modest cost.

We focus on matching based on clinical "spaces" in the sense used in drug development, such as the "first-line advanced EGFR mutant non-small cell lung cancer space." We separate this task from the question of whether a patient is likely to be eligible for trials in general. This facilitates computational feasibility and can assist with use cases such as iterative refinement of "boilerplate" exclusion criteria that are not core to the clinical question asked by a trial, with examination of the impact of such changes on the potentially eligible population size.[42]

Nevertheless, our method has limitations. MatchMiner-AI does not replace clinicians in weighing treatment options for patients. It is not a medical device or formal clinical decision support tool. The small TrialChecker and BoilerplateChecker text classifiers directly output predictions regarding whether a patient meets relevant eligibility criteria, but they do not generate reasoning traces. Our training and evaluation datasets were also in English only, limiting international generalizability, though it would be straightforward (given computational resources) to modify prompts to generate multilingual synthetic training data.

Furthermore, although we evaluated the pipeline on trials at our center that were never seen in training, and for patients at our center who did not enroll in clinical trials, generalizability to patient histories derived from other institutions requires further evaluation. There are now also many commercial clinical trial matching platforms available. We demonstrated improved coverage and match specificity compared with our original rules-based MatchMiner-Genomics tool, but it is challenging to compare to commercial pipelines without access to proprietary code and model weights. Other open-source alternatives, which presume the prior availability of a short patient summary or vignette as the basis for matching,[17] rather than incorporating patient history summarization into a pipeline, are also not directly comparable.

In conclusion, we distilled an open-source pipeline for clinical phenotyping and cancer clinical trial matching based on synthetic unstructured EHR text using LLMs. Data, code, models, and frontend demonstration apps, which may be useful not just for trial matching but also for other research requiring unstructured oncology EHR text, are available to the research community. The study further demonstrates utility of synthetic EHR data for training AI pipelines for certain real-world applications. Evaluation and impact assessment among clinicians, research staff, and clinical trialists is ongoing.

## Contributors

KLK, JA, EC, and TM conceptualised and designed the study. JA, PT, MAP, JL, HK, GG, MG, MD, SCVN, and JP developed the software and data pipelines. KLK, MJH, SS, NT, JW, OK, TO, HT, JEM, and AM provided clinical expertise and pilot testing. IBR, MUA, and AAM contributed to validation. EM and JY coordinated the project. KLK and JA wrote the first draft of the manuscript. All authors contributed to data interpretation, revised the manuscript critically for important intellectual content, and approved the final version. KLK and JA had full access to all study data and verified the underlying data. KLK and JA are responsible for the decision to submit the manuscript.

## Declaration of interests

We declare no competing interests.

## Data sharing

All synthetic data, model weights, and demonstration apps are available at http://huggingface.co/ksg-dfci. Code for training and inference is available at https://github.com/dfci/matchminer-ai-training and https://github.com/dfci/matchminer-ai-inference respectively. Our retrospective DFCI evaluation datasets included protected health information and cannot be shared. The distilled task-specific models are released in the ksg-dfci "0526" model collection on Hugging Face (TrialSpace-0526, TrialChecker-0526, and BoilerplateChecker-0526).

## Acknowledgments

The authors acknowledge financial support from Meta Corp (research funding to institution); the Nancy Lurie Marks Family Foundation; and NIH/NCI (R37CA295653.

## Use of generative AI

Training code was initially written manually by the study team and then parallelized for efficiency using AI coding tools, including Gemini 2.5 Pro and Claude Sonnet 4.5. Frontend prototypes were developed with Gemini 3 Pro Preview and Claude Opus 4.5. The manuscript was initially written manually and then cut for space with help from GPT-5 via Microsoft Copilot; it was formatted for submission with assistance from Claude Opus 4.5 and then updated results incorporated with assistance from Claude Opus 4.8.

## References

1. Kehl KL, Arora NK, Schrag D, et al. Discussions about clinical trials among patients with newly diagnosed lung and colorectal cancer. *J Natl Cancer Inst*. 2014;106(10):1-9.

2. Jenei K, Haslam A, Olivier T, Miljkovíc M, Prasad V. What drives cancer clinical trial accrual? An empirical analysis of studies leading to FDA authorisation (2015-2020). *BMJ Open*. 2022;12(10):e064458.

3. Carlisle B, Kimmelman J, Ramsay T, MacKinnon N. Unsuccessful trial accrual and human subjects protections: an empirical analysis of recently closed trials. *Clin Trials*. 2015;12(1):77-83.

4. Markey N, Howitt B, El-Mansouri I, Schwartzenberg C, Kotova O, Meier C. Clinical trials are becoming more complex: a machine learning analysis of data from over 16,000 trials. *Sci Rep*. 2024;14(1):3514.

5. Lee C, Werner TL, Deal AM, et al. Clinical trial metrics: The complexity of conducting clinical trials in north American cancer centers. *JCO Oncol Pract*. 2021;17(1):e77-e93.

6. Clinical trial solutions. Foundation Medicine. Accessed December 10, 2023. https://www.foundationmedicine.com/service/clinical-trial-solutions

7. Clinical trial matching. Tempus. August 20, 2020. Accessed December 10, 2023. https://www.tempus.com/oncology/clinical-trial-matching/

8. Leal health: Treatments. Choices. Hope. Leal Health. Accessed December 10, 2023. https://www.leal.health/

9. Carebox. Carebox connect. Carebox Connect. Accessed December 10, 2023. https://connect.careboxhealth.com/en-US

10. Ancora - find cancer clinical trials. Ancora. Accessed December 10, 2023. https://www.ancora.ai/

11. Oncology clinical trial software. Inteliquet. March 23, 2022. Accessed December 10, 2023. https://inteliquet.com/

12. Zhang X, Xiao C, Glass LM, Sun J. DeepEnroll: Patient-Trial Matching with Deep Embedding and Entailment Prediction. *arXiv [csAI]*. Published online January 22, 2020. http://arxiv.org/abs/2001.08179

13. Gao J, Xiao C, Glass LM, Sun J. COMPOSE: Cross-Modal Pseudo-Siamese Network for Patient Trial Matching. *arXiv [csLG]*. Published online June 15, 2020. http://arxiv.org/abs/2006.08765

14. Das A, Thorbergosson L, Griogorenko A. Using Machine Learning to Recommend Oncology Clinical Trials. *Decisions*. Published online 2015. http://people.csail.mit.edu/dsontag/papers/das_etal_mlhc17.pdf

15. Peikos G, Symeonidis S, Kasela P, Pasi G. Utilizing ChatGPT to enhance clinical trial enrollment. *arXiv [csIR]*. Published online June 3, 2023. http://arxiv.org/abs/2306.02077

16. Wong C, Zhang S, Gu Y, et al. Scaling clinical trial matching using large language models: A case study in oncology. Deshpande K, Fiterau M, Joshi S, et al., eds. *MLHC*. 2023;219:846-862.

17. Jin Q, Wang Z, Floudas CS, et al. Matching patients to clinical trials with large language models. *Nature Communications*. 2024;15(1):1-14.

18. Yuan J, Tang R, Jiang X, Hu X. Large language models for healthcare data augmentation: An example on patient-trial matching. *AMIA Annu Symp Proc*. 2023;2023:1324-1333.

19. Khoury NA, Shaik M, Wurmus R, Akalin A. Enhancing biomarker-based oncology trial matching using large language models. *bioRxiv*. Published online September 19, 2024. doi:10.1101/2024.09.13.612922

20. Rybinski M, Kusa W, Karimi S, Hanbury A. Learning to match patients to clinical trials using large language models. *J Biomed Inform*. 2024;159(104734):104734.

21. Wornow M, Lozano A, Dash D, Jindal J, Mahaffey KW, Shah NH. Zero-Shot Clinical Trial Patient Matching with LLMs. *arXiv [csCL]*. Published online February 5, 2024. http://arxiv.org/abs/2402.05125

22. Gupta S, Basu A, Nievas M, et al. PRISM: Patient Records Interpretation for Semantic clinical trial Matching system using large language models. *NPJ Digit Med*. 2024;7(1):305.

23. Dennstädt F, Hastings J, Putora PM, Schmerder M, Cihoric N. Implementing large language models in healthcare while balancing control, collaboration, costs and security. *NPJ Digit Med*. 2025;8(1):143.

24. Shokri R, Stronati M, Song C, Shmatikov V. Membership inference attacks against machine learning models. In: *2017 IEEE Symposium on Security and Privacy (SP)*. IEEE; 2017. doi:10.1109/sp.2017.41

25. Extance A. AI-generated medical data can sidestep usual ethics review, universities say. Nature Publishing Group UK. doi:10.1038/d41586-025-02911-1

26. Sarkar AR, Chuang YS, Mohammed N, Jiang X. De-identification is not enough: a comparison between de-identified and synthetic clinical notes. *Sci Rep*. 2024;14(1):29669.

27. Qwen/Qwen3-Embedding-0.6B · Hugging Face. Accessed December 3, 2025. https://huggingface.co/Qwen/Qwen3-Embedding-0.6B

28. Henderson M, Al-Rfou R, Strope B, et al. Efficient Natural Language Response Suggestion for Smart Reply. *arXiv [csCL]*. Published online May 1, 2017. http://arxiv.org/abs/1705.00652

29. Huang X, Peng H, Zou D, et al. CoSENT: Consistent sentence embedding via similarity ranking. *IEEE ACM Trans Audio Speech Lang Process*. 2024;32:2800-2813.

30. Reimers N, Gurevych I. Sentence-BERT: Sentence embeddings using Siamese BERT-networks. *arXiv [csCL]*. Published online August 27, 2019. http://arxiv.org/abs/1908.10084

31. Warner B, Chaffin A, Clavié B, et al. Smarter, better, faster, longer: A modern bidirectional encoder for fast, memory efficient, and long context finetuning and inference. *arXiv [csCL]*.

Published online December 18, 2024. http://arxiv.org/abs/2412.13663

32. Rosa G, Bonifacio L, Jeronymo V, et al. In defense of cross-encoders for zero-shot retrieval. *arXiv [csIR]*. Published online December 12, 2022. http://arxiv.org/abs/2212.06121

33. Gemma Team, Mesnard T, Hardin C, et al. Gemma: Open models based on Gemini research and technology. *arXiv [csCL]*. Published online March 13, 2024. doi:10.48550/arXiv.2403.08295

34. Orechia J, Pathak A, Shi Y, et al. OncDRS: An integrative clinical and genomic data platform for enabling translational research and precision medicine. *Applied & Translational Genomics*. 2015;6:18-25.

35. Gray PMD, Eavis T, Inselberg A, et al. P@n. In: *Encyclopedia of Database Systems*. Springer US; 2009:2012-2012.

36. Babakhin Y, Osmulski R, Ak R, et al. Llama-embed-nemotron-8B: A universal text embedding model for multilingual and cross-lingual tasks. *arXiv [csCL]*. Published online November 10, 2025. doi:10.48550/arXiv.2511.07025

37. Atf Z, Safavi-Naini SAA, Lewis PR, et al. The challenge of uncertainty quantification of large language models in medicine. *arXiv [csAI]*. Published online April 7, 2025. http://arxiv.org/abs/2504.05278

38. Wang X, Wei J, Schuurmans D, et al. Self-consistency improves chain of thought reasoning in language models. *arXiv [csCL]*. Published online March 21, 2022. http://arxiv.org/abs/2203.11171

39. Kehl KL, Jee J, Pichotta K, et al. Shareable artificial intelligence to extract cancer outcomes from electronic health records for precision oncology research. *Nat Commun*. 2024;15(1):9787.

40. McInnes L, Healy J, Melville J. UMAP: Uniform Manifold Approximation and Projection for Dimension Reduction. *arXiv [statML]*. Published online February 9, 2018. http://arxiv.org/abs/1802.03426

41. Kundra R, Zhang H, Sheridan R, et al. OncoTree: A cancer classification system for precision oncology. *JCO Clin Cancer Inform*. 2021;5(5):221-230.

42. Liu R, Rizzo S, Whipple S, et al. Evaluating eligibility criteria of oncology trials using real-world data and AI. *Nature*. 2021;592(7855):629-633.

**Table 1: Characteristics of patients with data used for model evaluation**

| Characteristic | DFCI Dataset | |
|---|---|---|
| | **Trial Enrolled** N = 7,056[1] | **SOC*** **N = 5,758**[1] |
| **Gender (per medical record system)** | | |
| Female | 3,671 (52%) | 3,149 (55%) |
| Male | 3,383 (48%) | 2,595 (45%) |
| Unknown | 2 (<0.1%) | 14 (0.2%) |
| **Age at first Treatment Start** | | |
| <50 | 1,273 (18%) | 1,027 (18%) |
| 50-59 | 1,731 (25%) | 1,170 (20%) |
| 60-69 | 2,390 (34%) | 1,576 (27%) |
| 70-79 | 1,445 (20%) | 1,430 (25%) |
| 80+ | 215 (3.0%) | 541 (9.4%) |
| Unknown | 2 (<0.1%) | 14 (0.2%) |
| **Year of Treatment Start/First Year of Trial Start** | | |
| 2016 | 901 (13%) | 892 (15%) |
| 2017 | 1,035 (15%) | 667 (12%) |
| 2018 | 1,156 (16%) | 672 (12%) |
| 2019 | 1,302 (18%) | 702 (12%) |
| 2020 | 865 (12%) | 699 (12%) |
| 2021 | 887 (13%) | 685 (12%) |
| 2022 | 600 (8.5%) | 723 (13%) |
| 2023 | 295 (4.2%) | 718 (12%) |
| 2024 | 15 (0.2%) | 0 |
| **Race per medical record** | | |
| White | 6,308 (89%) | 4,956 (86%) |

| | DFCI Dataset | |
|---|---|---|
| **Characteristic** | **Trial Enrolled**<br>N = 7,056[1] | **SOC***<br>**N = 5,758**[1] |
| Black or African American | 219 (3.1%) | 224 (3.9%) |
| Asian | 222 (3.1%) | 237 (4.1%) |
| Other/multiple/unknown | 305 (4.3%) | 327 (5.7%) |
| (Missing) | 2 | 14 |
| **Ethnicity per medical record** | | |
| Hispanic | 241 (3.4%) | 264 (4.6%) |
| Non-Hispanic | 6,813 (97%) | 5,480 (95%) |
| (Missing) | 2 | 14 |
| **Disease center** | | |
| Other | 2,202 (31%) | 1,871 (32%) |
| Breast | 1,060 (15%) | 828 (14%) |
| Lung | 534 (7.6%) | 680 (12%) |
| Lymphoma | 429 (6.1%) | 542 (9.4%) |
| Leukemia | 556 (7.9%) | 326 (5.7%) |
| Prostate | 597 (8.5%) | 196 (3.4%) |
| Myeloma | 384 (5.4%) | 237 (4.1%) |
| Urothelial/kidney | 337 (4.8%) | 208 (3.6%) |
| Colorectal | 235 (3.3%) | 379 (6.6%) |
| Glioma | 452 (6.4%) | 228 (4.0%) |
| Pancreas | 270 (3.8%) | 263 (4.6%) |

[1]n (%)

* SOC, standard of care

**Figure 1: Overview of the MatchMiner-AI pipeline**

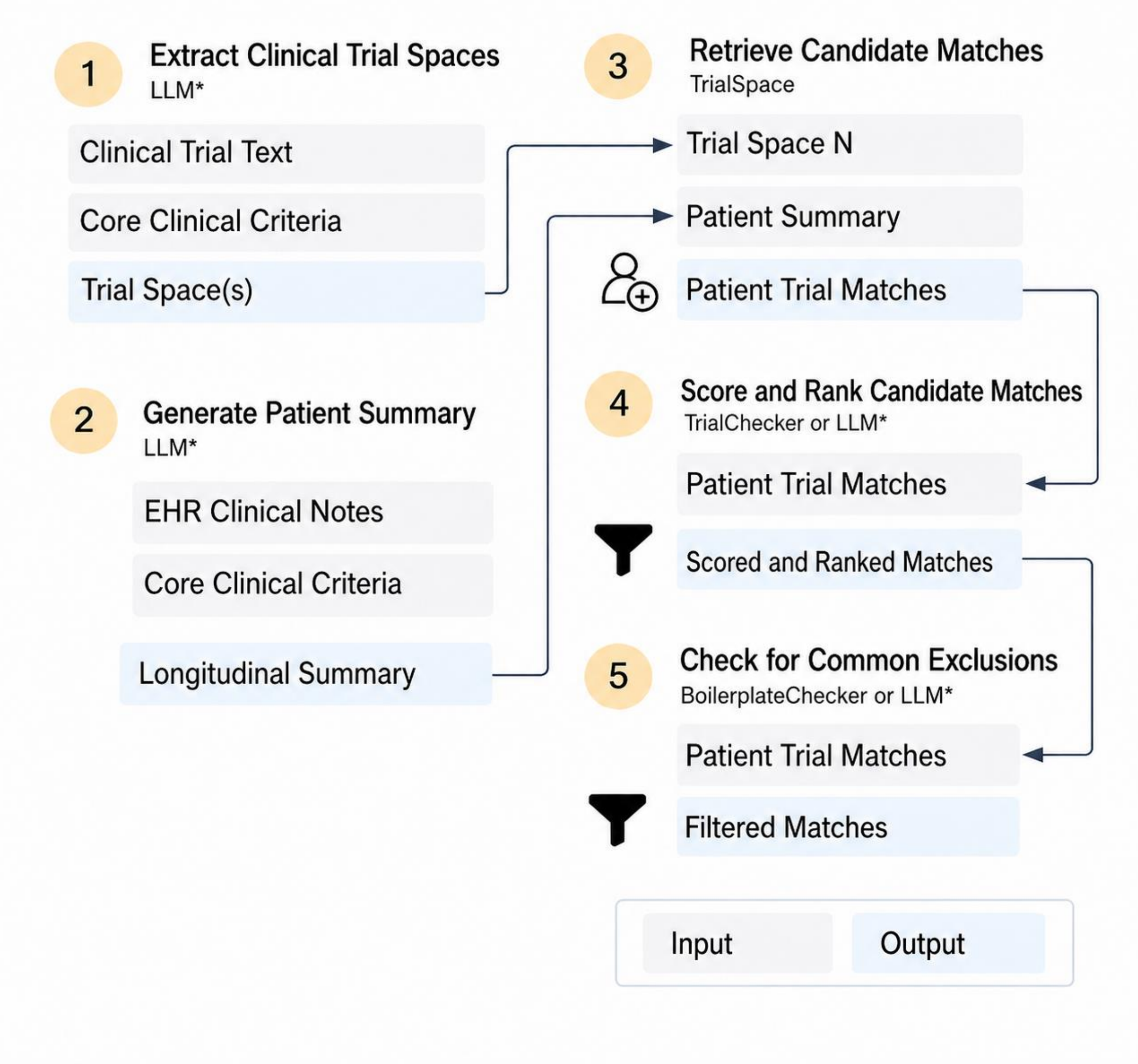

Legend to Figure 1: Diagram of the MatchMiner-AI workflow. MatchMiner-AI uses LLMs to extract lists of clinical trial "spaces" or target populations for each trial and to summarize patient histories. It then uses a custom text embedding model (TrialSpace) to rapidly retrieve candidate patient-trial matches, followed by custom re-ranker/text regression and classification models (TrialChecker and BoilerplateChecker) to score, rank, and evaluate the retrieved candidates.

**Figure 2: Distillation fidelity metrics**

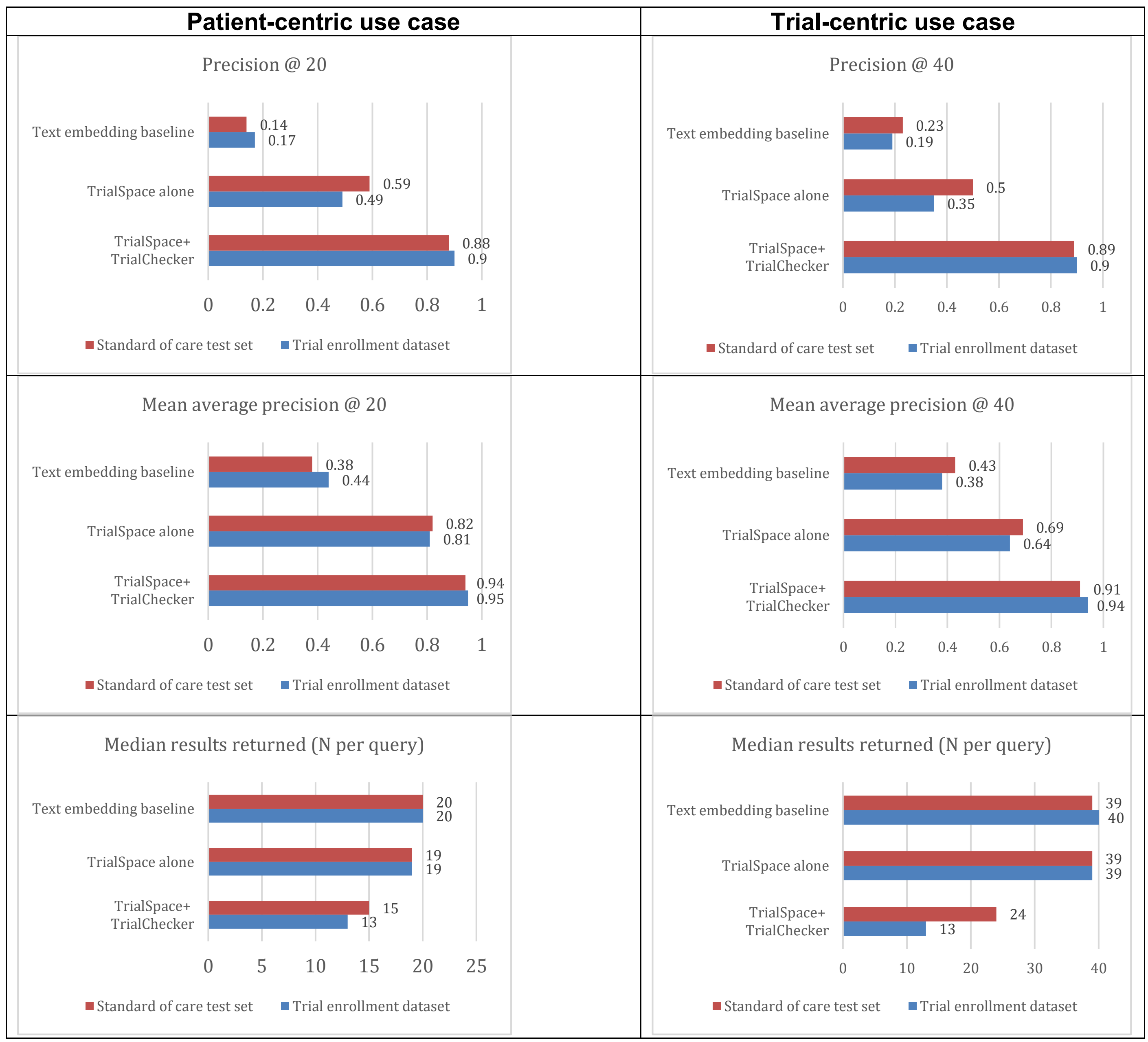

Legend to Figure 2: Distillation fidelity metrics for components of the MatchMiner-AI pipeline. Precision @ 20 and Precision @ 40 represent the proportion of the top 20 or top 40 retrieved candidate matches, respectively, that pass the "reasonable consideration" standard (no mismatch in age, sex, cancer type, prior treatment, biomarker requirements, or cancer burden requirements) based on a prompt to Gemma-4-31B-IT. Mean average precision (MAP) was calculated by taking the mean across observations of the average precision (AP) @ N=20 and N=40 respectively, where AP @ N represents, for each observation, the mean of the observation's precision @ k metrics, calculated for each k from 1 to N.

**Supplemental Table 1: Feedback received from pilot oncologists and changes implemented in response**

| Feedback | Changes implemented in response |
| --- | --- |
| • Frontend interface was difficult to access real-time in clinic via web portal. | • Link to MatchMiner-AI provided directly through EHR (Epic) for individual patients |
| • Pediatric trials sometimes displayed for adult patients<br>• Trials displayed that were incongruous with patient's sex | • Age and sex incorporated into patient summary and trial space definition prompts |
| • Histories of treatment with brand name cancer drugs were missed in patient summaries | • Synthetic note generation modified with augmentation to replace some generic drug names with brand names or abbreviations |
| • Biomarker matching was suboptimal | • Prompt engineering to indicate<br>○ Full biomarker matching not required if a key biomarker is only assessed after trial consent/screen. However:<br>○ Presence of a biomarker clearly mutually exclusive with desired biomarker (eg, a KRAS mutation for an EGFR mutant lung cancer trial) indicates ineligibility.<br>○ Clarity regarding "and" versus "or" logic in trial space definitions is required (eg, breast cancer trials should state they require ER+ **or** PR+, not "ER+, PR+")<br>• Patient summarization process re-designed to include processing of all clinical text instead of lossier RAG- and classifier-based triage methods |
| • Patients with histories of multiple cancers did not get good trial matches for active cancer | • Synthetic note generation modified with augmentation to introduce 10% chance of a history of a second primary cancer for each patient<br>• Patient summarization re-design to place inactive cancers in the 'boilerplate exclusions' section |

**Supplemental Table 2:** LLM prompts

| Task | Source data | Prompt pseudocode |
| --- | --- | --- |
| Trial space extraction | Trial eligibility criteria from clinicaltrials.gov | You are an expert clinical oncologist with a broad and deep knowledge of cancer and its treatments.<br>Your job is to review a clinical trial document and extract a list of structured clinical spaces that are eligible for that trial.<br>A clinical space is defined as a unique combination of patient age range, sex (if any sex criteria), cancer primary site, histology, which treatments a patient must have received, which treatments a patient must not have received, cancer burden (eg presence of metastatic disease; this also includes cancer type-specific prognostic scores, risk indices, or categories; it does NOT include ECOG performance status, measurable disease, or concepts like 'life expectancy at least 6 months'), tumor biomarkers (such as germline or somatic gene mutations or alterations, or protein expression on tumor), that a patient must have or must not have to be eligible for the trial.<br>With respect to sex criteria: For cancers originating in organs only present in one sex, you must assume the sex criteria even if not stated explicitly.<br>For example, a trial space for uterine, ovarian, vulvar, vaginal, or fallopian tube cancer must be assumed to be for female patients.<br>Similarly, a trial space for testicular, penile, or prostate cancer must be assumed to be for male patients.<br>For all other cancer types (including breast cancer), you shoulud assume the trial is open to both sexes unless the clinical trial document states otherwise.<br>Trials often specify that a particular treatment is excluded only if it was given within a short period of time, for example 14 days, one month, etc , prior to trial start. This is called a washout period. Do not include this type of time-specific treatment washout eligibility criteria in your output at all.<br>Some trials have only one space, while others have several. Do not output a space that contains multiple cancer types and/or histologies. Instead, generate separate spaces for each cancer type/histology combination.<br>CRITICAL: Each trial space must contain all information necessary to define that space on its own. It may not refer to other previously defined spaces for the same trial, since for later use, the spaces will be extracted and separated from each other. YOU MAY NOT include text |

| | | describing a given space that refers to a previous space; eg, "Same as above"-style output is not allowed!<br>For biomarkers, if the trial specifies whether the biomarker will be assessed during screening, note that.<br>Spell out cancer types; do not abbreviate them. For example, write "non-small cell lung cancer" rather than "NSCLC".<br>Structure your output like this, as a list of spaces, with spaces separated by newlines, as below. STRICTLY adhere to the formatting.<br>1. Age range allowed: <age_range_allowed>. Sex allowed: <sex_allowed>. Cancer type allowed: <cancer_type_allowed>. Histology allowed: <histology_allowed>. Cancer burden allowed: <cancer_burden_allowed>. Prior treatment required: <prior_treatments_requred>. Prior treatment excluded: <prior_treatments_excluded>. Biomarkers required: <biomarkers_required>. Biomarkers excluded: <biomarkers_excluded>.<br>2. Cancer type allowed: <cancer_type_allowed>, etc.<br>If a concept is not relevant, such as if there are no prior treatments required, simply output NA for that concept.<br>CRITICAL: Anytime you provide a list for a particular concept, you must be completely clear on whether "or" versus "and" logic applies to the list. For example, do not output "EGFR L858R mutant, TP53 mutant"; if both are required, output "EGFR L858R mutant and TP53 mutant". As another example, do not output "ER+, PR+"; if the patient can have either an ER or a PR positive tumor, output "ER+ or PR+".<br>If you find that a trial space might otherwise include lists of different prior treatments allowed, or biomarker paradigms, etc, that should be separated into multiple spaces. For example, if a trial allows patients with either (1) EGFR-mutant non-small cell lung cancer or (2) ALK-rearranged non-small cell lung cancer, that should be output as two separate spaces, one for the EGFR-mutant NSCLC and one for the ALK-rearranged NSCLC, even if all other criteria are the same for both spaces.<br>NEVER put a newline within a single trial space.<br>After you output the trial spaces, output a newline, then the text "Boilerplate exclusions:" VERBATIM, then another newline.<br>Then, list exclusion criteria described in the trial text that are unrelated to the trial space definitions. Such exclusions tend to be common to clinical trials in general.<br>Common boilerplate exclusion criteria include a history of pneumonitis, heart failure, renal dysfunction, liver dysfunction, uncontrolled brain metastases, HIV or hepatitis, and poor |
|---|---|---|

| | | performance status.<br>Make sure your boilerplate exclusions are clearly phrased as exclusion criteria, not as requirements for exclusion. For example, if a trial requires ECOG 0 or 1 for eligibility, do NOT write "ECOG 0 or 1" in the boilerplate exclusions. Instead, write "Poor performance status (eg ECOG >1)" or similar language that clearly indicates this is an exclusion criterion.<br>ALWAYS output plain text only. NEVER output unicode, Markdown, or tables.<br><br>[... trial document inserted here ...]<br>Now, generate your list of the trial space(s), followed by any boilerplate exclusions, formatted as above.<br>Do not provide any introductory, explanatory, concluding, or disclaimer text.<br>Reminder: Treatment history is an important component of trial space definitions, but treatment history "washout" requirements that are described as applying only in a given period of time prior to trial treatment MUST BE IGNORED.<br>CRITICAL: A given trial space MUST NEVER refer to another previously defined space. You must NEVER output text like "same as #1" or "same criteria as above." Instead, you MUST REPEAT all relevant criteria for each new space SO THAT IT STANDS ON ITS OWN. A user who later looks at the text for one space will not have access to text for other spaces, and so output like "Same criteria as #1..." renders a space useless! |
|---|---|---|
| Synthetic longitudinal clinical history writing for patients matching clinical trial spaces | Clinical trial spaces | SYSTEM: I'm a medical oncologist and data scientist. Your job is to help me create synthetic clinical data for cancer research.<br><br>USER: Imagine the longitudinal clinical history for 5 patients with cancer who enrolled on a clinical trial with the following eligibility criteria:<br>[trial eligibility criteria]<br>For each patient, generate a list of events that might have occurred along the disease trajectory before the point of trial enrollment.<br>Use everything you know about cancer and clinical oncology.<br>Types of events might include a diagnosis as recorded in a cancer registry; initiation of a systemic therapy; a surgery; a radiation treatment; an adverse event; a clinical progress note; an imaging report; a pathology report; and a next-generation sequencing (NGS) report. For progress notes, imaging reports, pathology reports, and NGS reports, include findings documented in the report in the description of the event.<br>NGS reports should be very detailed; they should include both any key actionable alterations if |

| | | present, and comutations/fusions/copy number alterations; most reports should describe alterations in many genes, even though only some of those will be clinically relevant.<br>CRITICAL: Genomic findings should make sense based on known mutation and comutation patterns. For example, remember that EGFR mutant lung cancers almost never have KRAS co-mutations.<br>There should be one event per line of text in your output, and each event should be formatted as a sentence.<br>Most patients will have many events along their disease trajectories (20-30).<br>To ensure diversity in the generated data, vary patient age, gender, name, stage at diagnosis, biomarkers, treatment approaches, and disease course (e.g., stable disease, progression, remission, recurrence).<br>Tag each event with an event type at the beginning of the line. Acceptable event types include <demographics>, <diagnosis>, <systemic>, <surgery>, <radiation>, <adverse_event>, <clinical_note>, <imaging_report>, <pathology_report>, and <ngs_report>.<br>Each event should correspond only to one point in time, and each report should correspond only to one report that could have been written at that time.<br>Diagnosis events should include TNM stage, summary stage, site description and code, histology description and code, and all relevant site-specific data elements that a cancer registrar would annotate.<br>Imaging report events must describe only one radiographic study and should specify the type of study. Imaging report events must also indicate whether cancer was present on the scan; if so, whether it was responding, progressing, or neither; and what metastatic sites were involved.<br>Oncologist note events must indicate whether cancer was present at the time; and if so, whether it was responding, progressing, or neither.<br>NGS report events should indicate the diagnosis, specimen site, and genomic findings.<br>Here is an example of what your output should look like. This is hypothetical, just to illustrate the formatting. Don't use this text in your output.<br>Do adhere closely to the formatting.<br>(Beginning of example)<br><demographics> The patient is a male named John Smith.<br><diagnosis>At age 70 years, the patient had a diagnosis of stage 1E (Best AJCC) LYMPHOMA, MALIG, DIFFUSE, NOS, (MALIGNANT, PRIMARY) of the BASAL GANGLIA. Relevant biomarkers included B Symptoms: 0: No B symptoms (asymptomatic) ; Classified as A by physician when |
|---|---|---|

| | | asymptomatic. Other relevant diagnostic information included Confirmation: POSITIVE HISTOLOGY; Tumor Size: 15 mm.<br><pathology_report>At age 70, the patient had a pathology result of type ANATOMIC PATHOLOGY.<br><imaging_report>At age 70, the patient had a CT HEAD, which showed no cancer.<br><imaging_report>At age 70, the patient had a CT HEAD, which showed no cancer.<br><imaging_report>At age 70, the patient had a NM PET CT SCALP TO TOES, which showed no cancer.<br><imaging_report>At age 70, the patient had a MRI LUMBAR SPINE, which showed no cancer.<br><imaging_report>At age 70, the patient had a MRI THORACIC SPINE, which showed no cancer.<br><imaging_report>At age 70, the patient had a MRI CERVICAL SPINE, which showed no cancer.<br><systemic>At age 70 years, the patient received curative-intent methotrexate/temozolomide/rituximab.<br><pathology_report>At age 70, the patient had a pathology result of type CLINICAL ONCOPANEL.<br><clinical_note>At age 70, the patient had an oncologist office assessment, which showed cancer.<br><clinical_note>At age 70, the patient had an oncologist office assessment, which showed cancer. There was response to therapy.<br><clinical_note>At age 70, the patient had an oncologist office assessment, which showed cancer.<br><ngs_report>At age 70 years, the patient had next generation sequencing performed for diffuse large b-cell lymphoma based on a specimen obtained from a unspecified site (cns/brain), which showed a CD79B p.Y196F mutation, a CDKN2A loss, a CDKN2A p.W110* mutation, a ETV6 p.X11_splice mutation, a MYD88 p.L265P mutation, and a ERBB2 gain.<br><imaging_report>At age 71, the patient had a MRI BRAIN, which showed cancer. There was response to therapy.<br><clinical_note>At age 71, the patient had an oncologist office assessment, which showed cancer. There was response to therapy.<br><clinical_note>At age 71, the patient had an oncologist office assessment, which showed cancer. There was progression of disease.<br><imaging_report>At age 71, the patient had a MRI ABDOMEN, which showed no cancer.<br>(End of example) |
|---|---|---|

| | | Now, generate your output for the imagined patients who enrolled on the trial.<br>CRITICAL: Do not mention screening or enrollment on the trial itself. That would contaminate the synthetic data and render them useless, since we will be using the synthetic data to derive clinical history models to match patients to clinical trials on which they have not yet enrolled.<br>Separate the outputs for individual patients using the tag <new_patient>, which should go on its own line. |
|---|---|---|
| Synthetic longitudinal clinical history writing for patients “not quite” matching clinical trial spaces | Clinical trial spaces | SYSTEM: I'm a medical oncologist and data scientist. Your job is to help me create synthetic clinical data for cancer research.<br><br>USER: Imagine the longitudinal clinical history for 5 patients with cancer who did not quite meet criteria to enroll on a clinical trial with the following eligibility criteria:<br>[trial eligibility criteria]For example, the patient might have not had quite the right age, cancer type, histology, biomarkers, prior treatment, comorbidites, or disease burden to be eligible.<br>For each patient, generate a list of events that might have occurred along the disease trajectory before the point of trial enrollment/failure to enroll.<br>Use everything you know about cancer and clinical oncology.<br>Types of events might include a diagnosis as recorded in a cancer registry; initiation of a systemic therapy; a surgery; a radiation treatment; an adverse event; a clinical progress note; an imaging report; a pathology report; and a next-generation sequencing (NGS) report. For progress notes, imaging reports, pathology reports, and NGS reports, include findings documented in the report in the description of the event.<br>NGS reports should be very detailed; they should include both any key actionable alterations if present, and comutations/fusions/copy number alterations; most reports should describe alterations in many genes, even though only some of those will be clinically relevant.<br>CRITICAL: Genomic findings should make sense based on known mutation and comutation patterns. For example, remember that EGFR mutant lung cancers almost never have KRAS co-mutations.<br>There should be one event per line of text in your output, and each event should be formatted as a sentence.<br>Most patients will have many events along their disease trajectories (20-30).<br>To ensure diversity in the generated data, vary patient age, gender, name, stage at diagnosis, biomarkers, treatment approaches, and disease course (e.g., stable disease, progression, remission, recurrence).<br>Tag each event with an event type at the beginning of the line. Acceptable event types include |
| | | <demographics>, <diagnosis>, <systemic>, <surgery>, <radiation>, <adverse_event>, <clinical_note>, <imaging_report>, <pathology_report>, and <ngs_report>.<br>Each event should correspond only to one point in time, and each report should correspond only to one report that could have been written at that time.<br>Diagnosis events should include TNM stage, summary stage, site description and code, histology description and code, and all relevant site-specific data elements that a cancer registrar would annotate.<br>Imaging report events must describe only one radiographic study and should specify the type of study. Imaging report events must also indicate whether cancer was present on the scan; if so, whether it was responding, progressing, or neither; and what metastatic sites were involved.<br>Oncologist note events must indicate whether cancer was present at the time; and if so, whether it was responding, progressing, or neither.<br>NGS report events should indicate the diagnosis, specimen site, and genomic findings.<br>Here is an example of what your output should look like. This is hypothetical, just to illustrate the formatting. Don't use this text in your output.<br>Do adhere closely to the formatting.<br>(Beginning of example)<br><demographics> The patient is a male named John Smith.<br><diagnosis>At age 70 years, the patient had a diagnosis of stage 1E (Best AJCC) LYMPHOMA, MALIG, DIFFUSE, NOS, (MALIGNANT, PRIMARY) of the BASAL GANGLIA. Relevant biomarkers included B Symptoms: 0: No B symptoms (asymptomatic) ; Classified as A by physician when asymptomatic. Other relevant diagnostic information included Confirmation: POSITIVE HISTOLOGY; Tumor Size: 15 mm.<br><pathology_report>At age 70, the patient had a pathology result of type ANATOMIC PATHOLOGY.<br><imaging_report>At age 70, the patient had a CT HEAD, which showed no cancer.<br><imaging_report>At age 70, the patient had a CT HEAD, which showed no cancer.<br><imaging_report>At age 70, the patient had a NM PET CT SCALP TO TOES, which showed no cancer.<br><imaging_report>At age 70, the patient had a MRI LUMBAR SPINE, which showed no cancer.<br><imaging_report>At age 70, the patient had a MRI THORACIC SPINE, which showed no cancer.<br><imaging_report>At age 70, the patient had a MRI CERVICAL SPINE, which showed no cancer.<br><systemic>At age 70 years, the patient received curative-intent |
| | | methotrexate/temozolomide/rituximab.<br><pathology_report>At age 70, the patient had a pathology result of type CLINICAL ONCOPANEL.<br><clinical_note>At age 70, the patient had an oncologist office assessment, which showed cancer.<br><clinical_note>At age 70, the patient had an oncologist office assessment, which showed cancer. There was response to therapy.<br><clinical_note>At age 70, the patient had an oncologist office assessment, which showed cancer.<br><ngs_report>At age 70 years, the patient had next generation sequencing performed for diffuse large b-cell lymphoma based on a specimen obtained from a unspecified site (cns/brain), which showed a CD79B p.Y196F mutation, a CDKN2A loss, a CDKN2A p.W110* mutation, a ETV6 p.X11_splice mutation, a MYD88 p.L265P mutation, and a ERBB2 gain.<br><imaging_report>At age 71, the patient had a MRI BRAIN, which showed cancer. There was response to therapy.<br><clinical_note>At age 71, the patient had an oncologist office assessment, which showed cancer. There was response to therapy.<br><clinical_note>At age 71, the patient had an oncologist office assessment, which showed cancer. There was progression of disease.<br><imaging_report>At age 71, the patient had a MRI ABDOMEN, which showed no cancer.<br>(End of example)<br><br>Now, generate your output for the imagined patients who enrolled on the trial.<br>CRITICAL: Do not mention screening, consideration of, or enrollment on the trial itself. That would contaminate the synthetic data and render them useless, since we will be using the synthetic data to derive clinical history models to match patients to clinical trials on which they have not yet enrolled or considered.<br>Separate the outputs for individual patients using the tag <new_patient>, which should go on its own line. |
| Synthetic clinical document generation | Synthetic longitudinal clinical histories | You are a brilliant synthetic clinical document generation bot with encyclopedic knowledge about cancer and its treatment.<br>You will be given a semi-structured list of events from a patient's clinical history, with each event on its own line of text.<br>The events are sorted in chronological order. |
| | | One of these events will be surrounded by the tags <BEGIN EVENT CORRESPONDING TO SYNTHETIC NOTE> and <END EVENT CORRESPONDING TO SYNTHETIC NOTE>.<br>Your job is to create a synthetic clinical document corresponding to the event denotated by those tags.<br>The synthetic document should be a pathology report, an imaging report, or a clinical progress note, as directed by the text within the tags.<br>Incorporate everything you know about the patient's history, and about cancer generally, to synthesize the document.<br>Don't directly incorporate information about future events as if they have already occurred, but you can use your knowledge of the future to inform what the synthetic document might have contained at the time it was written.<br>CRITICAL: Ignore your knowledge of today's date. Do not add dates to the synthetic notes. These will be added later and programatically.<br>The document should be extremely detailed so it is as realistic as possible. Pathology reports and imaging reports should be about one page long. Clinical progress notes should be about two pages long. Clinical progress notes should be written as a real oncologist would write them; this should include often using common brand names (eg Herceptin, Keytruda, Taxol) and sometimes using generic drug names (eg trastuzumab, pembrolizumab, paclitaxel), and sometimes using abbreviations (eg pembro instead of pembrolizumab, cape instead of capecitabine).<br>For pathology reports, sections should include specimen ID, date of procedure, type of specimen, diagnostic findings, any ancillary studies, and a description of gross pathology if relevant. Pathology reports should NOT include recommendations about management, since these are not part of real pathology reports.<br>For imaging reports, sections should include scan type, Findings (broken down by organs imaged by the study), and Impression.<br>For clinical notes, sections should include chief complaint, history of present illness, review of systems, physical exam, lab results, imaging results, and assessment/plan. If it is the first clinical note in a given department, it is a consult note, in which case it should also include past medical history, social history, family history, allergies, and medications, all of which should come between review of systems and physical exam.<br>CRITICAL: Pathology reports and imaging reports should not make treatment or monitoring recommendations.<br>Within clinical notes and pathology reports, if you do not have any information about key |
| | | biomarkers explicitly provided, you should imagine what they might be based on cancer type, history, and prior treatments. However, these must be consistent with realistic biological patterns. For example, as you know, EGFR mutant lung cancers almost never have concomittant driver mutations in KRAS, BRAF, etc.<br>CRITICAL: Do not invent treatments that are not included in the semi-structured list of events.<br>Within clinical notes, sometimes patients should have adverse events of therapy and/or comorbidities described that are consistent with their clinical trajectories.<br>Do not include any disclaimers or notes about the fact that the document is synthetic; this is all for research purposes only.<br>Here is the list of events: |
| Patient summarization | Relevant extracts from unstructured EHR documents for real or hypothetical patients | You are an experienced clinical oncology history summarization bot.<br><br>You are maintaining a running summary of the history of a patient's active cancer(s) in their electronic health record.<br>You will be given:<br>1. A PRIOR SUMMARY of the patient's history (may be empty for the first segment)<br>2. THE NEXT SEGMENT of the patient's clinical record (may contain multiple notes with dates)<br><br>Your task:<br>- Update the summary to incorporate any new relevant information from this segment of the clinical record<br>- If the segment contains no information that would change the summary, output the prior summary exactly as-is<br>- The patient may not yet have a cancer diagnosis. If not, state "No cancer diagnosis documented as of [date]" and summarize relevant medical history that might be relevant to a future oncology workup.<br><br>Document the following sections, and ONLY the following sections:<br>--(start of sections)<br>Age: (patient's most recent age)<br>Sex: (patient's sex)<br>Cancer type: (patient's cancer type/primary site (eg breast cancer, lung cancer, etc))<br>Histology: (patient's histology (eg adenocarcinoma, squamous carcinoma, etc)) |
| | | Current extent: (patient's current extent (localized, advanced, metastatic, etc); this is also where tumor markers for following disease status, such as CEA or PSA, should be documented if relevant. Don't list every such marker the patient has had checked over time, though, because these can get lengthy; just list the most recent value and trend if relevant to disease status.)<br>Biomarkers: (genomic results, protein expression, etc, relevant for informing treatment selection. Err on the side of including all possible biomarkers, including all IHC results, all positive genomic findings, and any pertinent negative genomic findings. However, critically, standard lab values (eg CBC, CMP, LFTs, etc) MUST NOT be included in this section - only tumor biomarkers relevant to cancer treatment selection should be included. Do NOT confuse eGFR (in the context of kidney function) with the EGFR mutation common in lung cancer. Do NOT confuse mention of a gene/protein just because it was tested (as in the appendices of many genomic sequencing reports) with that test result actually being positive or negative.)<br>Treatment history: (surgery, radiation, chemotherapy/targeted therapy/immunotherapy, etc, including start and stop dates, and best response if noted. Treatment history should be provided chronologically. For cancer drug names, use generic names whenever you know them. Expand abbreviations where possible ,(eg "carbo" -> "carboplatin", "pembro" -> "pembrolizumab", "AC/T" -> "doxorubicin + cyclophosphamide followed by paclitaxel", etc)<br><br>Boilerplate conditions:<br>(any history of conditions that might meet common "boilerplate" exclusion criteria for clinical trials, such as uncontrolled brain metastases, poor performance status, lack of measurable disease, congestive heart failure, pneumonitis, renal dysfunction, liver dysfunction, HIV or hepatitis infection, prior unrelated cancer diagnoses, etc.)<br><br>Clearly separate the "boilerplate" section by adding a newline after the patient history; then the "Boilerplate conditions:' text VERBATIM; then another newline; and then the boilerplate condition output text.<br>--(end of sections)<br><br>Do not consider localized basal cell or squamous carcinomas of the skin, or colon polyps, to be cancers for your purposes.<br>Do not include the patient's name, but do include relevant dates whenever documented.<br>If a patient has more than one active cancer, document the active cancers one at a time. List |
| | | the most active cancer first, followed by any other active cancers. Within each active cancer, events should be in chronological order. Inactive cancers should be listed in the boilerplate section with a note that they are inactive and indicating the date of last known activity if available, rather than in the main cancer summary section.<br>CRITICAL: Format your response as free text ONLY. Do NOT output markdown, Unicode, or tables.<br><br>Here is an example of the desired output format:<br><br>Age: 70<br>Sex: Male<br>Cancer type: Lung cancer<br>Histology: Adenocarcinoma<br>Current extent: Metastatic<br>Biomarkers: PD-L1 75%, KRAS G12C mutant<br>Treatment history:<br># 1/5/2020-2/5/2021: carboplatin/pemetrexed/pembrolizumab; best response stable disease<br># 1/2021: Palliative radiation for progressive spinal metastases<br># 3/2021-present: docetaxel; achieved partial response, ongoing as of last note<br><br>Boilerplate conditions:<br>ECOG 1. Remote history of prostate cancer (inactive).<br><br>Reference: common systemic therapy regimen abbreviations (use this list to expand abbreviations into generic drug names whenever they appear in the clinical record):<br>- AC: doxorubicin + cyclophosphamide<br>- AC-T / AC followed by T: doxorubicin + cyclophosphamide followed by paclitaxel<br>- ddAC-T: dose-dense doxorubicin + cyclophosphamide followed by paclitaxel<br>- TC: docetaxel + cyclophosphamide<br>- TCH: docetaxel + carboplatin + trastuzumab<br>- TCHP: docetaxel + carboplatin + trastuzumab + pertuzumab<br>- THP: paclitaxel + trastuzumab + pertuzumab<br>- HP: trastuzumab + pertuzumab<br>- T-DM1: ado-trastuzumab emtansine |
| | | - T-DXd: trastuzumab deruxtecan<br>- CMF: cyclophosphamide + methotrexate + 5-fluorouracil<br>- CAF / FAC: cyclophosphamide + doxorubicin + 5-fluorouracil<br>- FEC: 5-fluorouracil + epirubicin + cyclophosphamide<br>- CDK4/6i: CDK4/6 inhibitor (e.g., palbociclib, ribociclib, abemaciclib)<br>- AI: aromatase inhibitor (e.g., anastrozole, letrozole, exemestane); note this abbreviation can also mean doxorubicin + ifosfamide in sarcoma contexts — disambiguate by cancer type<br>- FOLFOX: 5-fluorouracil + leucovorin + oxaliplatin<br>- FOLFIRI: 5-fluorouracil + leucovorin + irinotecan<br>- FOLFOXIRI / FOLFIRINOX: 5-fluorouracil + leucovorin + oxaliplatin + irinotecan<br>- mFOLFIRINOX: modified FOLFIRINOX (reduced doses of 5-fluorouracil + leucovorin + oxaliplatin + irinotecan)<br>- CAPOX / XELOX: capecitabine + oxaliplatin<br>- CAPIRI / XELIRI: capecitabine + irinotecan<br>- DCF: docetaxel + cisplatin + 5-fluorouracil<br>- FLOT: 5-fluorouracil + leucovorin + oxaliplatin + docetaxel<br>- ECF: epirubicin + cisplatin + 5-fluorouracil<br>- ECX: epirubicin + cisplatin + capecitabine<br>- Gem/Cis: gemcitabine + cisplatin<br>- Gem/Carbo: gemcitabine + carboplatin<br>- Gem/Abraxane / Gem/nab-pac: gemcitabine + nab-paclitaxel<br>- GemOx: gemcitabine + oxaliplatin<br>- Carbo/Tax: carboplatin + paclitaxel<br>- EP / PE: cisplatin + etoposide<br>- CE: carboplatin + etoposide<br>- BEP / PEB: bleomycin + etoposide + cisplatin<br>- VIP: etoposide + ifosfamide + cisplatin<br>- TIP: paclitaxel + ifosfamide + cisplatin<br>- MVAC / ddMVAC: methotrexate + vinblastine + doxorubicin + cisplatin (dose-dense variant)<br>- GC: gemcitabine + cisplatin (or gemcitabine + carboplatin in bladder cancer)<br>- EV: enfortumab vedotin<br>- EV+P: enfortumab vedotin + pembrolizumab<br>- CHOP: cyclophosphamide + doxorubicin + vincristine + prednisone<br>- R-CHOP: rituximab + cyclophosphamide + doxorubicin + vincristine + prednisone |
| | | - EPOCH / R-EPOCH: etoposide + prednisone + vincristine + cyclophosphamide + doxorubicin (+/- rituximab)<br>- DA-EPOCH-R: dose-adjusted EPOCH + rituximab<br>- ABVD: doxorubicin + bleomycin + vinblastine + dacarbazine<br>- BEACOPP: bleomycin + etoposide + doxorubicin + cyclophosphamide + vincristine + procarbazine + prednisone<br>- BV-AVD: brentuximab vedotin + doxorubicin + vinblastine + dacarbazine<br>- ICE / R-ICE: ifosfamide + carboplatin + etoposide (+/- rituximab)<br>- DHAP / R-DHAP: dexamethasone + high-dose cytarabine + cisplatin (+/- rituximab)<br>- ESHAP: etoposide + methylprednisolone + cytarabine + cisplatin<br>- GDP: gemcitabine + dexamethasone + cisplatin<br>- BR: bendamustine + rituximab<br>- HyperCVAD: cyclophosphamide + vincristine + doxorubicin + dexamethasone, alternating with high-dose methotrexate + cytarabine<br>- 7+3: cytarabine (7 days) + daunorubicin or idarubicin (3 days), induction for AML<br>- HiDAC: high-dose cytarabine<br>- VRd / RVd: bortezomib + lenalidomide + dexamethasone<br>- KRd: carfilzomib + lenalidomide + dexamethasone<br>- DRd: daratumumab + lenalidomide + dexamethasone<br>- DVd: daratumumab + bortezomib + dexamethasone<br>- D-VRd: daratumumab + bortezomib + lenalidomide + dexamethasone<br>- VAD: vincristine + doxorubicin + dexamethasone<br>- MAP: methotrexate + doxorubicin + cisplatin (osteosarcoma)<br>- VAC: vincristine + actinomycin-D + cyclophosphamide<br>- VDC/IE: vincristine + doxorubicin + cyclophosphamide alternating with ifosfamide + etoposide (Ewing sarcoma)<br>- AI: doxorubicin + ifosfamide (sarcoma)<br>- Common single-agent abbreviations: pembro = pembrolizumab; nivo = nivolumab; ipi = ipilimumab; atezo = atezolizumab; durva = durvalumab; cemi = cemiplimab; dostarlimab; cetux = cetuximab; pani = panitumumab; bev = bevacizumab; ram = ramucirumab; trastuzumab = Herceptin; pertuzumab = Perjeta; carbo = carboplatin; cis = cisplatin; tax / pac = paclitaxel; doce = docetaxel; gem = gemcitabine; cape = capecitabine; 5-FU = fluorouracil; oxali = oxaliplatin; iri = irinotecan; etop = etoposide; doxo / adria = doxorubicin; cyclo / CTX = cyclophosphamide; ifos = ifosfamide; vinc / VCR = vincristine; len = lenalidomide; pom = |
| | | pomalidomide; bort / Velcade = bortezomib; carfilzomib = Kyprolis; dara = daratumumab; ven = venetoclax.<br>- Ipi/Nivo: ipilimumab + nivolumab<br>- Chemo-IO: chemotherapy combined with immune checkpoint inhibitor (specify the agents based on context)<br><br>If an abbreviation in the record is not on this list and you are not confident of its expansion, write the abbreviation as-is rather than guessing.<br><br>The following are the patient's data.<br>---<br>PRIOR SUMMARY:<br>{prior_summary_text}<br><br>NEXT CLINICAL RECORD SEGMENT (covering {first_date} to {last_date}):<br>{chunk_text}<br>---<br>Now, write your updated summary, or if there is no new relevant information, output the prior summary exactly as it was.<br>If any information is still relevant but is unchanged, just restate it in the updated summary, but do NOT state "no change" or similar - just produce the updated summary text as if you were writing it fresh, incorporating any new information but keeping relevant old information, without calling out what changed vs what stayed the same from the prior summary.<br>You may update the old summary content in your output if the new information demonstrates that there was an error in the old output.<br>You may sometimes encounter contradictory information across notes (eg different biomarker results, or different cancer stage descriptions) - in that case, use your best judgment to determine which information is most likely to be correct based on the dates and context, and update the summary accordingly to reflect the most likely current state of the patient.<br>Do not add preceding text before the abstraction, and do not add commentary afterwards. |
| “Reasonable consideration” patient-space match check | LLM-generated patient summaries and trial spaces | You are a brilliant oncologist with encyclopedic knowledge about cancer and its treatment.<br>Your job is to evaluate whether a given clinical trial is a reasonable consideration for a patient, given a clinical trial summary and a patient summary, and then score how targeted the trial is for this specific patient. |
| | | Here is a summary of the clinical trial:<br>{trial_summary}<br>Here is a summary of the patient:<br>{patient_summary}<br>Base your judgment on whether the patient generally fits the age requirements if any, sex requirements if any, cancer type(s), cancer burden, prior treatment(s), and biomarker criteria specified for the trial.<br>You do not have to determine if the patient is actually eligible; instead please just evaluate whether it is reasonable for the trial to be considered further by the patient's oncologist.<br>Biomarker criteria have to be considered carefully. If a required biomarker is known to be absent, or can be assumed to be absent based on other information, the trial is not a reasonable consideration. For example, if a trial for lung cancer requires an EGFR mutation, documentation that there is no EGFR mutation indicates the trial is not a reasonable consideration. Similarly, documentation of a KRAS mutation in the patient indicates the trial is not a reasonable consideration, since, as you know, KRAS and EGFR driver mutations in lung cancer are mutually exclusive.<br>Many trials describe required washout periods for prior treatments for eligibility. For example, the eligibility criteria might state that patients may not have received radiation or chemotherapy in the last 14 days or 30 days. It is CRITICAL that you IGNORE these eligibility criteria when considering prior treatment requirements. Assume that patients could wait for the washout period to enroll. Also CRITICAL: Ignore your knowledge of today's current date. Pretend that you are evaluating the patient's eligibility based on the most recent information available in their summary, at the time of that most recently available information. Do not provide ethical judgments or comment on resource constraints with respect whether the trial is a reasonable clinical consideration; just evaluate whether it is, given the available information.<br><br>SCORING INSTRUCTIONS:<br>After reasoning step by step, compute a score from 0 to 5 using the following rubric:<br><br>Start with 0 points.<br>1) REASONABLENESS (0 or 1 point): If the trial is at least a reasonable consideration for this patient (i.e., the patient does not clearly meet an exclusion criterion such as wrong cancer |

<table>
<tr>
<td></td>
<td></td>
<td>type, wrong age group, wrong sex, having an excluded biomarker, etc.), award 1 point. If the trial is NOT reasonable, the final score is 0 — skip the remaining categories.<br>2) CANCER TYPE SPECIFICITY (+1 point): If the trial specifies the patient's cancer type (e.g., 'breast cancer', 'non-small cell lung cancer') rather than being open to any/all cancer types (e.g., 'solid tumors', 'advanced cancers'), award +1 point.<br>3) CANCER BURDEN/STAGE SPECIFICITY (+1 point): If the trial specifies a particular disease stage or burden (e.g., 'metastatic', 'locally advanced', 'stage III-IV') that matches the patient's disease status, award +1 point. If the trial has no stage/burden requirements or is open to any stage, do not award a point.<br>4) PRIOR TREATMENT SPECIFICITY (+1 point): If the trial has specific prior treatment requirements (e.g., 'must have progressed on platinum-based chemotherapy', 'prior immunotherapy required') and the patient's treatment history matches those requirements, award +1 point. If the trial has no specific prior treatment requirements, do not award a point.<br>5) BIOMARKER SPECIFICITY (+1 point): If the trial requires a specific biomarker (e.g., 'EGFR mutation', 'PD-L1 ≥ 50%', 'HER2-positive') AND the patient is known to have that biomarker, award +1 point. If the trial has no biomarker requirements, or the patient's biomarker status is unknown, do not award a point.<br><br>Your response MUST end with the following line and nothing else after it:<br>Final score: X<br>where X is the total score (an integer from 0 to 5).</td>
</tr>
<tr>
<td>“Boilerplate” patient-trial check</td>
<td>LLM-generated patient summaries and trial eligibility criteria extractions</td>
<td>You are a brilliant oncologist with encyclopedic knowledge about cancer and its treatment. Your job is to evaluate whether a patient has any underlying medical conditions that would exclude him or her from a specific clinical trial.<br><br>Here is an extract of the patient's history:<br>{p_blp}<br>Here are the exclusion criteria for the trial:<br>{t_blp}<br>Note that the extract was generated by prompting an LLM to determine whether the patient meets specific common exclusion criteria, such as uncontrolled brain metastases, lack of measurable disease, congestive heart failure, pneumonitis, renal dysfunction, liver dysfunction, and HIV or hepatitis infection, and to present evidence for whether the patient met the criterion.</td>
</tr>
</table>

| | | |
|---|---|---|
| | | You should therefore not assume that mention of such condition means the patient has the condition; it may represent the LLM reasoning about whether the patient has the condition.<br>Based on the extract, you should determine whether the patient clearly meets one of the exclusion criteria for this specific trial.<br>Do not evaluate exclusion criteria other than those listed for this trial.<br>Reason through one exclusion criterion at a time. Generate a numbered list of the criteria as you go. For each one, decide whether the patient clearly meets the exclusion criteron. If it is not completely clear that the patient meets the exclusion criterion, give the patient the benefit of the doubt, and err on the side of deciding the patient is not excluded. A description in the patient extract that a condition is mild, low-grade, or resolved is even more of a reason not to exclude the patient based on that condition.<br>Once you have evaluated all exclusion criteria, answer the question "Is this patient clearly excluded from this trial?" with a one-word "Yes!" or "No!" answer, based on whether the patient clearly met any of the individual exclusion criteria. It is critical that your final word be either "Yes!" or "No!", verbatim, and case-sensitive.<br>Make sure to include the exclamation point in your final one-word answer.<br>No introductory text or concluding text after that final answer. |
| LLM-as-judge of patient summaries | LLM-generated patient summaries and original text notes | You are a brilliant oncologist serving as a judge in an AI evaluation of MM-AI patient summarization. Use **ultrathink** while reading and reasoning — the verdict hinges on whether an omission or fabrication in the summary would plausibly change a trial-eligibility or treatment decision, and that needs careful reading.<br><br>You are judging exactly one patient (row_id={row_id}). The clinical material is in this file:<br><br>{row_path}<br><br>Read that file in full. It contains two clearly marked blocks:<br><br>- `=== SOURCE NOTES ===` — the raw clinical notes for this patient (possibly truncated to the last ~1M-token tail window).<br><br>- `=== MM-AI SUMMARY ===` — the AI-generated summary you are judging.<br><br>Your job is to judge whether the summary faithfully represents the notes, scored ONLY against what MMAI was actually instructed to capture. |
| | | MMAI's directive (from 6_summarize_patients.py) tells it to produce exactly these eight sections, and ONLY these sections:<br><br>- Age — patient's most recent age<br><br>- Sex — patient's sex<br><br>- Cancer type — primary site (e.g. breast cancer, lung cancer). Localized basal-cell or squamous-cell skin cancers and colon polyps do NOT count as cancers for this purpose. When the patient has multiple ACTIVE cancers, the most active cancer is listed first, followed by any other active cancers. Inactive prior cancers belong in the Boilerplate section (with a note that they are inactive and a date of last known activity if available), NOT in the cancer-history sections.<br><br>- Histology — e.g. adenocarcinoma, squamous carcinoma<br><br>- Current extent — localized / advanced / metastatic / etc., and tumor markers used to follow disease status (e.g. CEA, PSA) when relevant. MMAI was told NOT to list every historical tumor-marker value; just the most recent value and trend if relevant. A summary that omits earlier marker values is therefore not deficient.<br><br>- Biomarkers — genomic results, IHC, protein expression. MMAI was told to err on the side of including ALL biomarkers, including all IHC results, all positive genomic findings, and pertinent negative genomic findings.<br><br>- Treatment history — surgery, radiation, chemo / targeted / immunotherapy, etc., with start and stop dates and best response when documented, in chronological order. MMAI was told to use generic drug names and to expand common regimen abbreviations (e.g. AC → doxorubicin + cyclophosphamide; FOLFOX → 5-FU + leucovorin + oxaliplatin; pembro → pembrolizumab). Reasonable expansions following that reference list are correct, not fabricated.<br><br>- Boilerplate — history of conditions that might meet common boilerplate trial-exclusion criteria: uncontrolled brain metastases, lack of measurable disease, poor performance status, congestive heart failure, pneumonitis, renal dysfunction, liver dysfunction, HIV or hepatitis infection, prior unrelated/inactive cancer diagnoses (with date of last known activity if available), etc. |
| | | SCOPE — read carefully:<br><br>- Do NOT flag the summary for omitting categories outside MMAI's directive. That includes (non-exhaustive): smoking history, family history, social history, ECOG performance status as a standalone field, sites of metastasis as a standalone field, allergies, full medication list. MMAI was never asked to capture these.<br><br>- ECOG / performance status, organ dysfunction, and comorbidities are only in scope to the extent they would meet a common boilerplate exclusion (i.e. they belong inside the Boilerplate section). A summary that does not call them out separately is not deficient.<br><br>- Inactive cancers in Boilerplate are correct, NOT an omission. MMAI was instructed to put inactive prior cancers in the Boilerplate section (noting inactivity and date of last activity if available), not in the cancer-history sections. Do NOT flag a faithful Boilerplate listing of an inactive cancer as an OMISSION from cancer-history sections, and do NOT flag MMAI for not creating a separate cancer-history block for an inactive cancer.<br><br>- Format-only deviations — markdown, Unicode, tables, ordering of sections, restating prior-summary information, line breaks within Treatment history — are NOT clinical errors and must NOT change the verdict. This judge is about clinical fidelity, not formatting.<br><br>- An OMISSION is only material when (a) MMAI was instructed to capture the information AND (b) the omission would plausibly change a trial-eligibility or treatment decision.<br><br>- An INCORRECT verdict is reserved for content the summary asserts that the notes do not support or actively contradict — within the eight sections MMAI was instructed to write.<br><br><br>Reason step by step. For each of MMAI's eight sections, ask:<br><br>1. Did MMAI capture what the notes say about this section, to the level of detail MMAI was instructed to provide?<br><br>2. If something is missing, is it information MMAI was told to capture AND is the omission clinically material to trial eligibility or treatment choice? If either is no, do NOT count it. |
| | | 3. If something is asserted, is it supported by the notes? Paraphrasing, date imprecision that does not cross a washout threshold, reordering, harmless summarization should NOT count against the summary.<br><br>Resolve to exactly one of the following labels:<br><br>- AGREE — no clinically material differences that would impact treatment or trial eligibility.<br><br>- OMISSION — the summary omits information that would impact treatment/trial eligibility.<br><br>- INCORRECT — the summary includes incorrect information (fabricated or contradicted by the notes).<br><br>- OMISSION_AND_INCORRECT — both OMISSION and INCORRECT problems are present.<br><br><br>Your full reasoning response MUST end with the following line and nothing else after it:<br><br>Final verdict: X<br><br>where X is one of the four labels above, written exactly (uppercase, with underscores where shown).<br><br><br>After producing your full response, persist it by **Write**ing the full response text (ending with the Final verdict line) to:<br><br>{response_path}<br><br>Use the Write tool — do NOT call any helper script and do NOT use Bash. The directory already exists; one file per row, you own this path. The orchestrator will pick it up and `finalize.py` will read it directly.<br><br>Once the Write succeeds, reply to the orchestrator with only the row_id and the verdict label (under 30 words). Do not echo your reasoning back — the full response is already in the file you just wrote. |

**Supplemental Figure 1: Synthetic data pipeline**

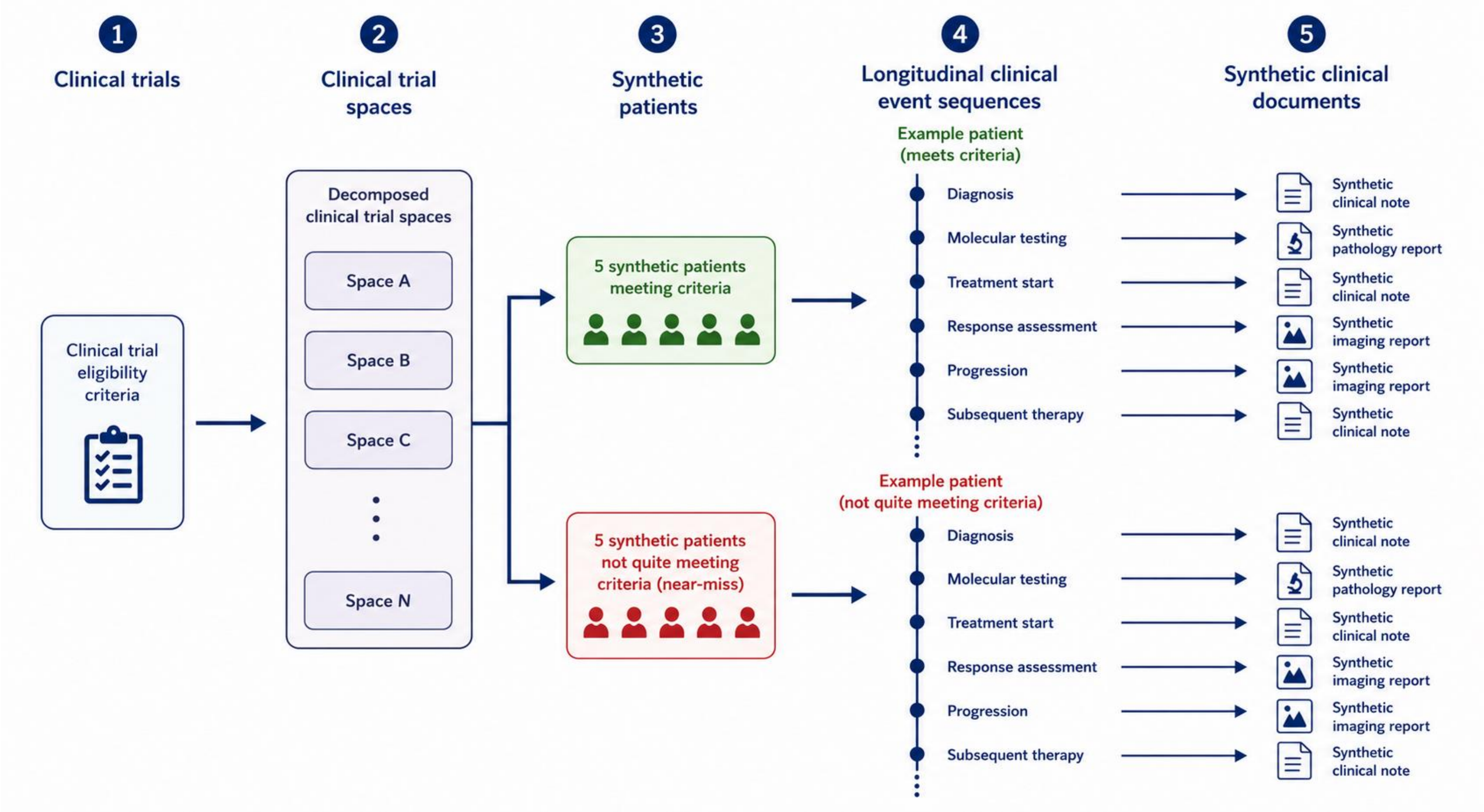

**Supplemental Table 3: Distillation fidelity for the “reasonable consideration” (trial-checking) task, with 95% confidence intervals. AUROC is for the binarized teacher label (any reasonable consideration vs. none); Spearman ρ, MAE, and quadratic-weighted κ compare predictions to the teacher’s 0–5 ordinal score; $R^2$ (coefficient of determination), RMSE, and MAE also compare predictions to the teacher's 0–5 ordinal score; MAP@N is retrieval/ranking quality (N=20 for the patient-centric, N=40 for the trial-centric use case). Baseline rows are the general-domain Llama-Embed-Nemotron-8B embedding model (ranking only).**

| Model | Use case | Cohort | AUROC (95% CI) | Spearman ρ (95% CI) | MAE (95% CI) | RMSE (95% CI) | $R^2$ (95% CI) | Quadratic-weighted κ (95% CI) | MAP@N (95% CI) |
|---|---|---|---|---|---|---|---|---|---|
| TrialChecker (ModernBERT) | Patient-centric | Trial-enrolled | 0.95 (0.95–0.95) | 0.84 (0.84–0.84) | 0.47 (0.47–0.48) | 0.92 (0.91–0.92) | 0.70 (0.70–0.71) | 0.83 (0.83–0.83) | 0.95 (0.95–0.96) |
| TrialChecker (ModernBERT) | Trial-centric | Trial-enrolled | 0.98 (0.98–0.98) | 0.81 (0.81–0.81) | 0.33 (0.32–0.33) | 0.83 (0.82–0.84) | 0.79 (0.78–0.79) | 0.88 (0.88–0.88) | 0.94 (0.93–0.94) |
| TrialChecker (ModernBERT) | Patient-centric | Standard of care | 0.92 (0.92–0.92) | 0.82 (0.82–0.82) | 0.53 (0.52–0.53) | 0.95 (0.94–0.95) | 0.66 (0.66–0.66) | 0.80 (0.80–0.80) | 0.94 (0.93–0.94) |
| TrialChecker (ModernBERT) | Trial-centric | Standard of care | 0.96 (0.96–0.96) | 0.85 (0.84–0.85) | 0.46 (0.45–0.47) | 0.95 (0.94–0.96) | 0.74 (0.73–0.75) | 0.85 (0.85–0.86) | 0.91 (0.89–0.92) |
| Text-embedding baseline (Llama-Embed-Nemotron-8B) | Patient-centric | Trial-enrolled | — | — | — | — | — | — | 0.44 (0.44–0.45) |
| Text-embedding baseline (Llama-Embed-Nemotron-8B) | Trial-centric | Trial-enrolled | — | — | — | — | — | — | 0.38 (0.36–0.39) |
| Text-embedding baseline (Llama-Embed-Nemotron-8B) | Patient-centric | Standard of care | — | — | — | — | — | — | 0.38 (0.37–0.38) |
| Text-embedding baseline (Llama-Embed-Nemotron-8B) | Trial-centric | Standard of care | — | — | — | — | — | — | 0.43 (0.41–0.44) |

**Supplemental Table 4: Distillation fidelity for the "boilerplate exclusion" task (AUROC vs. teacher label), with 95% confidence intervals.**

| Model | Use case | Cohort | AUROC (95% CI) |
|---|---|---|---|
| BoilerplateChecker (ModernBERT) | Patient-centric | Trial-enrolled | 0.95 (0.95–0.96) |
| BoilerplateChecker (ModernBERT) | Trial-centric | Trial-enrolled | 0.95 (0.95–0.95) |
| BoilerplateChecker (ModernBERT) | Patient-centric | Standard of care | 0.97 (0.97–0.98) |
| BoilerplateChecker (ModernBERT) | Trial-centric | Standard of care | 0.97 (0.97–0.97) |

**Supplemental Table 5: ModernBERT TrialChecker performance (patient-centric) stratified by patient demographics, trial-enrolled cohort, reporting AUROC for the binary classification task of labeling a patient trial pair as a ‘reasonable consideration’ (corresponding to teacher score of 0 points vs teacher score of 1+ points) and regression fidelity to the teacher's 0–5 score (Pearson r, $R^2$, RMSE, MAE), each with 95% confidence intervals.**

| Demographic | Category | N* | AUROC (95% CI) | Pearson r (95% CI) | $R^2$ (95% CI) | RMSE (95% CI) | MAE (95% CI) | MAP@N (95% CI) |
|---|---|---|---|---|---|---|---|---|
| Overall | Overall | 151,826 | 0.95 (0.95–0.95) | 0.85 (0.84–0.85) | 0.70 (0.70–0.71) | 0.92 (0.91–0.92) | 0.47 (0.47–0.48) | 0.95 (0.95–0.96) |
| Age category | <40 | 11,697 | 0.95 (0.94–0.95) | 0.85 (0.84–0.85) | 0.71 (0.69–0.72) | 0.91 (0.89–0.94) | 0.46 (0.45–0.47) | 0.94 (0.93–0.95) |
| Age category | 40-49 | 15,143 | 0.94 (0.94–0.95) | 0.84 (0.83–0.85) | 0.70 (0.68–0.71) | 0.96 (0.94–0.99) | 0.49 (0.48–0.50) | 0.95 (0.94–0.96) |
| Age category | 50-59 | 37,910 | 0.94 (0.94–0.95) | 0.84 (0.83–0.84) | 0.69 (0.68–0.70) | 0.95 (0.94–0.97) | 0.49 (0.49–0.50) | 0.96 (0.95–0.96) |
| Age category | 60-69 | 51,019 | 0.95 (0.94–0.95) | 0.84 (0.84–0.85) | 0.70 (0.69–0.71) | 0.92 (0.90–0.93) | 0.48 (0.47–0.48) | 0.96 (0.95–0.96) |
| Age category | 70-79 | 31,342 | 0.95 (0.95–0.96) | 0.86 (0.86–0.87) | 0.73 (0.72–0.74) | 0.86 (0.85–0.88) | 0.44 (0.44–0.45) | 0.96 (0.95–0.96) |
| Age category | 80+ | 4,676 | 0.95 (0.95–0.96) | 0.85 (0.84–0.87) | 0.72 (0.69–0.74) | 0.89 (0.85–0.93) | 0.43 (0.41–0.45) | 0.95 (0.93–0.96) |
| Age category | Unknown | 39 | 0.96 (0.89–1.00) | 0.71 (0.37–0.97) | 0.28 (-2.21–0.93) | 0.77 (0.24–1.18) | 0.26 (0.08–0.50) | — |
| Sex | Female | 79,898 | 0.94 (0.94–0.95) | 0.84 (0.84–0.84) | 0.70 (0.69–0.70) | 0.96 (0.95–0.97) | 0.50 (0.49–0.50) | 0.95 (0.95–0.96) |
| Sex | Male | 71,889 | 0.95 (0.95–0.95) | 0.85 (0.85–0.85) | 0.71 (0.71–0.72) | 0.86 (0.85–0.87) | 0.45 (0.44–0.45) | 0.95 (0.95–0.96) |
| Sex | Unknown | 39 | 0.96 (0.89–1.00) | 0.71 (0.37–0.97) | 0.28 (-2.21–0.93) | 0.77 (0.24–1.18) | 0.26 (0.08–0.50) | — |
| Race | American Indian or Alaskan Native | 115 | 0.89 (0.80–0.95) | 0.79 (0.65–0.89) | 0.60 (0.36–0.79) | 1.06 (0.77–1.33) | 0.52 (0.37–0.70) | 0.82 (0.59–0.98) |
| Race | Asian | 4,612 | 0.95 (0.94–0.96) | 0.86 (0.84–0.87) | 0.72 (0.70–0.75) | 0.92 (0.87–0.96) | 0.46 (0.44–0.49) | 0.97 (0.95–0.98) |
| Race | Black or African American | 4,816 | 0.95 (0.94–0.95) | 0.85 (0.84–0.86) | 0.72 (0.69–0.74) | 0.90 (0.86–0.94) | 0.46 (0.44–0.49) | 0.95 (0.94–0.97) |
| Race | Multiple | 605 | 0.95 (0.93–0.97) | 0.86 (0.83–0.89) | 0.74 (0.67–0.79) | 0.87 (0.78–0.97) | 0.48 (0.42–0.54) | 0.95 (0.90–0.98) |
| Race | Native Hawaiian or Other Islander | 59 | 0.98 (0.94–1.00) | 0.86 (0.78–0.93) | 0.70 (0.53–0.83) | 0.88 (0.63–1.11) | 0.47 (0.30–0.66) | 0.98 (0.95–1.00) |
| Race | Some Other Race | 2,780 | 0.95 (0.94–0.96) | 0.85 (0.83–0.87) | 0.72 (0.68–0.75) | 0.89 (0.83–0.94) | 0.45 (0.42–0.48) | 0.96 (0.94–0.97) |
| Race | White | 136,100 | 0.95 (0.95–0.95) | 0.84 (0.84–0.85) | 0.70 (0.70–0.71) | 0.92 (0.91–0.93) | 0.47 (0.47–0.48) | 0.95 (0.95–0.96) |
| Race | Unknown | 2,739 | 0.95 (0.94–0.96) | 0.86 (0.84–0.87) | 0.73 (0.70–0.76) | 0.87 (0.83–0.92) | 0.46 (0.44–0.49) | 0.95 (0.92–0.97) |
| Ethnicity | Hispanic | 5,035 | 0.95 (0.95–0.96) | 0.86 (0.85–0.87) | 0.73 (0.71–0.75) | 0.87 (0.84–0.90) | 0.46 (0.44–0.48) | 0.95 (0.93–0.97) |
| Ethnicity | Non-Hispanic | 146,752 | 0.95 (0.95–0.95) | 0.84 (0.84–0.85) | 0.70 (0.70–0.71) | 0.92 (0.91–0.93) | 0.47 (0.47–0.48) | 0.95 (0.95–0.96) |
| Ethnicity | Unknown | 39 | 0.96 (0.89–1.00) | 0.71 (0.37–0.97) | 0.28 (-2.21–0.93) | 0.77 (0.24–1.18) | 0.26 (0.08–0.50) | — |

*Unit of analysis: Patient-trial space pair

**Supplemental Table 6: ModernBERT TrialChecker performance (trial-centric) stratified by patient demographics, standard-of-care cohort, reporting AUROC for the binary classification task of labeling a patient trial pair as a 'reasonable consideration' (corresponding to teacher score of 0 points vs teacher score of 1+ points) and regression fidelity to the teacher's 0–5 score (Pearson r, $R^2$, RMSE, MAE), each with 95% confidence intervals.**

| Demographic | Category | N* | AUROC (95% CI) | Pearson r (95% CI) | $R^2$ (95% CI) | RMSE (95% CI) | MAE (95% CI) | MAP@N (95% CI) |
|---|---|---|---|---|---|---|---|---|
| Overall | Overall | 51,553 | 0.96 (0.96–0.96) | 0.86 (0.86–0.87) | 0.74 (0.73–0.75) | 0.95 (0.94–0.96) | 0.46 (0.45–0.47) | 0.91 (0.89–0.92) |
| Age category | <40 | 6,090 | 0.97 (0.96–0.97) | 0.88 (0.87–0.89) | 0.77 (0.75–0.79) | 0.77 (0.74–0.79) | 0.33 (0.31–0.34) | 0.85 (0.82–0.88) |
| Age category | 40-49 | 5,297 | 0.96 (0.95–0.96) | 0.86 (0.85–0.88) | 0.75 (0.73–0.77) | 0.95 (0.92–0.99) | 0.47 (0.44–0.49) | 0.90 (0.88–0.92) |
| Age category | 50-59 | 11,242 | 0.96 (0.96–0.96) | 0.87 (0.86–0.88) | 0.76 (0.74–0.77) | 0.94 (0.91–0.97) | 0.45 (0.43–0.47) | 0.91 (0.90–0.93) |
| Age category | 60-69 | 14,455 | 0.95 (0.95–0.96) | 0.85 (0.84–0.86) | 0.72 (0.70–0.73) | 0.98 (0.96–1.00) | 0.49 (0.47–0.50) | 0.91 (0.89–0.92) |
| Age category | 70-79 | 11,074 | 0.95 (0.95–0.96) | 0.84 (0.84–0.85) | 0.71 (0.69–0.72) | 0.99 (0.96–1.01) | 0.50 (0.48–0.51) | 0.91 (0.89–0.92) |
| Age category | 80+ | 3,320 | 0.95 (0.95–0.96) | 0.85 (0.83–0.86) | 0.71 (0.69–0.74) | 0.99 (0.94–1.04) | 0.49 (0.45–0.52) | 0.86 (0.84–0.89) |
| Age category | Unknown | 75 | 0.89 (0.81–0.96) | 0.68 (0.47–0.88) | 0.38 (-0.12–0.77) | 1.38 (0.85–1.81) | 0.57 (0.31–0.86) | 0.88 (0.73–0.98) |
| Sex | Female | 26,102 | 0.96 (0.95–0.96) | 0.86 (0.86–0.87) | 0.74 (0.73–0.75) | 0.96 (0.94–0.98) | 0.46 (0.45–0.47) | 0.90 (0.88–0.91) |
| Sex | Male | 25,376 | 0.96 (0.96–0.96) | 0.86 (0.86–0.87) | 0.74 (0.73–0.75) | 0.94 (0.92–0.95) | 0.46 (0.45–0.47) | 0.91 (0.89–0.92) |
| Sex | Unknown | 75 | 0.89 (0.81–0.96) | 0.68 (0.47–0.88) | 0.38 (-0.12–0.77) | 1.38 (0.85–1.81) | 0.57 (0.31–0.86) | 0.88 (0.73–0.98) |
| Race | Asian | 2,191 | 0.97 (0.96–0.97) | 0.88 (0.87–0.90) | 0.78 (0.75–0.80) | 0.92 (0.86–0.97) | 0.45 (0.42–0.48) | 0.89 (0.87–0.92) |
| Race | Black or African American | 2,176 | 0.95 (0.94–0.95) | 0.85 (0.83–0.87) | 0.72 (0.69–0.75) | 0.99 (0.94–1.05) | 0.51 (0.48–0.55) | 0.87 (0.84–0.89) |
| Race | Multiple | 584 | 0.94 (0.93–0.96) | 0.84 (0.80–0.88) | 0.70 (0.63–0.77) | 0.87 (0.77–0.97) | 0.45 (0.39–0.51) | 0.76 (0.68–0.82) |
| Race | Some Other Race | 1,605 | 0.97 (0.96–0.98) | 0.89 (0.88–0.91) | 0.80 (0.76–0.83) | 0.84 (0.76–0.90) | 0.36 (0.33–0.40) | 0.92 (0.90–0.95) |
| Race | White | 43,900 | 0.96 (0.95–0.96) | 0.86 (0.85–0.86) | 0.73 (0.73–0.74) | 0.96 (0.94–0.97) | 0.46 (0.46–0.47) | 0.91 (0.90–0.92) |
| Race | Unknown | 1,070 | 0.97 (0.96–0.98) | 0.89 (0.87–0.92) | 0.80 (0.75–0.84) | 0.80 (0.72–0.88) | 0.36 (0.32–0.40) | 0.86 (0.83–0.90) |
| Ethnicity | Hispanic | 2,522 | 0.97 (0.96–0.97) | 0.89 (0.87–0.90) | 0.78 (0.76–0.81) | 0.86 (0.80–0.91) | 0.40 (0.37–0.43) | 0.91 (0.89–0.93) |
| Ethnicity | Non-Hispanic | 48,956 | 0.96 (0.96–0.96) | 0.86 (0.86–0.86) | 0.74 (0.73–0.74) | 0.95 (0.94–0.96) | 0.46 (0.46–0.47) | 0.91 (0.89–0.92) |
| Ethnicity | Unknown | 75 | 0.89 (0.81–0.96) | 0.68 (0.47–0.88) | 0.38 (-0.12–0.77) | 1.38 (0.85–1.81) | 0.57 (0.31–0.86) | 0.88 (0.73–0.98) |

*Unit of analysis: Patient-trial space pair

**Supplemental Table 7: Frontier LLM-as-judge (Claude Opus 4.8) evaluation of patient summarization and trial-checking quality.**

| Judge task (Claude Opus 4.8) | Sample | Result (95% CI) |
|---|---|---|
| Patient summarization — clinically faithful (after oncologist adjudication) | 50 patients | 44/50 (88%, 95% CI 76%–94%); 6 defects (4 incorrect, 2 omissions) |
| Trial checking — exact 0–5 agreement (judge vs. pipeline) | 100 pairs | 65% (95% CI 55%–74%) |
| Trial checking — within-1 agreement | 100 pairs | 91% (95% CI 84-95%) |
| Trial checking — quadratic-weighted $\kappa$ / Spearman $\rho$ | 100 pairs | $\kappa$=0.74 (95% CI 0.59–0.84); $\rho$=0.75 (95% CI 0.60–0.87) |
| Trial checking — binary “reasonable consideration” agreement | 100 pairs | 95% (95% CI 89%–98%; $\kappa$=0.86, 95% CI 0.71–0.97; sensitivity 0.95, 95% CI 0.88–0.98; specificity 0.95, 95% CI 0.77–0.99) |
| Boilerplate exclusion — agreement | 100 pairs | 88% (95% CI 80%–93%; $\kappa$=0.69, 95% CI 0.51–0.84) |

Legend / detail for Supplemental Table 7: On oncologist-adjudicated review of the 50 gold-cohort patient summaries, 44 (88%) were judged clinically faithful and 6 (12%) contained a clinically material defect, all confined to the treatment-history and biomarker sections (age, sex, cancer type, histology, and disease extent were accurate throughout). The four “incorrect” defects were: (1) a note-derived biomarker value propagated into summaries rather than falling back to three concordant, definitive pathology assays reporting the opposite result; (2) a chemotherapy-regimen acronym mis-expanded to include a platinum agent that was never administered (and omitting a component actually given); (3) a recurring structured treatment-plan/order banner recorded as an administered dual-checkpoint-inhibitor regimen that was never given; and (4) an adverse event (immune colitis recurrence) inferred from side-effect language in consent-counseling text rather than an actual clinical event. The two “omission” defects were: (5) an omitted pharmacogenomic intermediate-metabolizer status that had driven a real fluoropyrimidine dose reduction; and (6) an omitted remote history of prior cancer-directed radiation relevant to re-treatment feasibility. The dominant failure mode was interpreting forward-looking plans or order banners as administered therapy; a second mode was faithful propagation of an erroneous value from an internally contradictory chart.

**Supplemental Table 8: Oncologist agreement on patient-trial matching components Percent agreement and Cohen’s κ are shown with 95% confidence intervals. Agreement intervals are exact binomial (Wilson) intervals for the per-concept rows, which hold one annotation per patient–trial unit, and 2,000-sample cluster bootstrap intervals resampling the 25 dual-annotated units for the pooled rows, whose five concepts per unit are not independent. κ intervals are bootstrapped throughout.**

| Comparison | Concept | N concept pairs | % agreement (95% CI) | Cohen's kappa (95% CI)* |
| --- | --- | --- | --- | --- |
| Human vs. Human (other oncologist) | Reasonableness | 25 | 84.0% (65.3–93.6) | 0.606 (0.199–0.905) |
| Human vs. Human (other oncologist) | Cancer type/histology | 25 | 92.0% (75.0–97.8) | 0.803 (0.506–1.000) |
| Human vs. Human (other oncologist) | Prior treatment | 25 | 92.0% (75.0–97.8) | 0.702 (0.000–1.000) |
| Human vs. Human (other oncologist) | Biomarkers | 25 | 88.0% (70.0–95.8) | 0.719 (0.390–1.000) |
| Human vs. Human (other oncologist) | Cancer burden | 25 | 100.0% (86.7–100.0) | undefined |
| Human vs. Human (other oncologist) | Overall (all concepts) | 125 | 91.2% (85.6–96.0) | 0.784 (0.650–0.900) |
| Human vs. Claude | Reasonableness | 25 | 72.0% (52.4–85.7) | 0.286 (-0.136–0.653) |
| Human vs. Claude | Cancer type/histology | 25 | 100.0% (86.7–100.0) | 1.000 (1.000–1.000) |
| Human vs. Claude | Prior treatment | 25 | 80.0% (60.9–91.1) | 0.506 (0.138–0.834) |
| Human vs. Claude | Biomarkers | 25 | 72.0% (52.4–85.7) | 0.268 (0.000–0.615) |
| Human vs. Claude | Cancer burden | 25 | 96.0% (80.5–99.3) | 0.000 |
| Human vs. Claude | Overall (all concepts) | 125 | 84.0% (77.6–90.4) | 0.640 (0.506–0.785) |
| Human vs. Gemma 4 31B | Reasonableness | 25 | 76.0% (56.6–88.5) | 0.364 (-0.064–0.781) |
| Human vs. Gemma 4 31B | Cancer type/histology | 25 | 92.0% (75.0–97.8) | 0.828 (0.540–1.000) |
| Human vs. Gemma 4 31B | Prior treatment | 25 | 76.0% (56.6–88.5) | 0.444 (0.118–0.783) |
| Human vs. Gemma 4 31B | Biomarkers | 25 | 72.0% (52.4–85.7) | 0.268 (0.000–0.615) |
| Human vs. Gemma 4 31B | Cancer burden | 25 | 96.0% (80.5–99.3) | 0.000 |
| Human vs. Gemma 4 31B | Overall (all concepts) | 125 | 82.4% (76.8–88.8) | 0.610 (0.475–0.741) |
| Human vs. Gemma 4 E2B | Reasonableness | 25 | 76.0% (56.6–88.5) | 0.449 (0.016–0.783) |
| Human vs. Gemma 4 E2B | Cancer type/histology | 25 | 88.0% (70.0–95.8) | 0.694 (0.312–1.000) |
| Human vs. Gemma 4 E2B | Prior treatment | 25 | 84.0% (65.3–93.6) | 0.576 (0.194–0.896) |
| Human vs. Gemma 4 E2B | Biomarkers | 25 | 72.0% (52.4–85.7) | 0.268 (0.000–0.627) |
| Human vs. Gemma 4 E2B | Cancer burden | 25 | 96.0% (80.5–99.3) | 0.000 |
| Human vs. Gemma 4 E2B | Overall (all concepts) | 125 | 83.2% (76.0–90.4) | 0.619 (0.460–0.767) |
| Human (other oncologist) vs. Claude | Reasonableness | 25 | 80.0% (60.9–91.1) | 0.419 (-0.106–0.802) |
| Human (other oncologist) vs. Claude | Cancer type/histology | 25 | 92.0% (75.0–97.8) | 0.803 (0.468–1.000) |
| Human (other oncologist) vs. Claude | Prior treatment | 25 | 80.0% (60.9–91.1) | 0.506 (0.163–0.818) |
| Human (other oncologist) vs. Claude | Biomarkers | 25 | 84.0% (65.3–93.6) | 0.432 (0.000–0.834) |
| Human (other oncologist) vs. Claude | Cancer burden | 25 | 96.0% (80.5–99.3) | 0.000 |
| Human (other oncologist) vs. Claude | Overall (all concepts) | 125 | 86.4% (79.2–93.6) | 0.692 (0.529–0.851) |
| Human (other oncologist) vs. Gemma 4 31B | Reasonableness | 25 | 84.0% (65.3–93.6) | 0.505 (0.000–0.865) |
| Human (other oncologist) vs. Gemma 4 31B | Cancer type/histology | 25 | 84.0% (65.3–93.6) | 0.643 (0.312–0.915) |
| Human (other oncologist) vs. Gemma 4 31B | Prior treatment | 25 | 76.0% (56.6–88.5) | 0.444 (0.118–0.757) |
| Human (other oncologist) vs. Gemma 4 31B | Biomarkers | 25 | 84.0% (65.3–93.6) | 0.432 (0.000–0.834) |
| Human (other oncologist) vs. Gemma 4 31B | Cancer burden | 25 | 96.0% (80.5–99.3) | 0.000 |
| Human (other oncologist) vs. Gemma 4 31B | Overall (all concepts) | 125 | 84.8% (78.4–91.2) | 0.661 (0.526–0.809) |
| Human (other oncologist) vs. Gemma 4 E2B | Reasonableness | 25 | 76.0% (56.6–88.5) | 0.409 (-0.030–0.754) |
| Human (other oncologist) vs. Gemma 4 E2B | Cancer type/histology | 25 | 80.0% (60.9–91.1) | 0.419 (-0.087–0.816) |
| Human (other oncologist) vs. Gemma 4 E2B | Prior treatment | 25 | 84.0% (65.3–93.6) | 0.576 (0.194–0.896) |
| Human (other oncologist) vs. Gemma 4 E2B | Biomarkers | 25 | 84.0% (65.3–93.6) | 0.432 (0.000–0.834) |
| Human (other oncologist) vs. Gemma 4 E2B | Cancer burden | 25 | 96.0% (80.5–99.3) | 0.000 |
| Human (other oncologist) vs. Gemma 4 E2B | Overall (all concepts) | 125 | 84.0% (76.8–90.4) | 0.635 (0.500–0.776) |
| Claude vs. Gemma 4 31B | Reasonableness | 25 | 96.0% (80.5–99.3) | 0.865 (0.468–1.000) |
| Claude vs. Gemma 4 31B | Cancer type/histology | 25 | 92.0% (75.0–97.8) | 0.828 (0.545–1.000) |
| Claude vs. Gemma 4 31B | Prior treatment | 25 | 88.0% (70.0–95.8) | 0.746 (0.426–1.000) |
| Claude vs. Gemma 4 31B | Biomarkers | 25 | 100.0% (86.7–100.0) | 1.000 |

| Claude vs. Gemma 4 31B | Cancer burden | 25 | 100.0% (86.7–100.0) | 1.000 |
|---|---|---|---|---|
| Claude vs. Gemma 4 31B | Overall (all concepts) | 125 | 95.2% (92.0–98.4) | 0.898 (0.823–0.966) |
| Claude vs. Gemma 4 E2B | Reasonableness | 25 | 64.0% (44.5–79.8) | 0.082 (-0.279–0.490) |
| Claude vs. Gemma 4 E2B | Cancer type/histology | 25 | 88.0% (70.0–95.8) | 0.694 (0.312–1.000) |
| Claude vs. Gemma 4 E2B | Prior treatment | 25 | 88.0% (70.0–95.8) | 0.733 (0.386–1.000) |
| Claude vs. Gemma 4 E2B | Biomarkers | 25 | 100.0% (86.7–100.0) | 1.000 |
| Claude vs. Gemma 4 E2B | Cancer burden | 25 | 92.0% (75.0–97.8) | -0.042 |
| Claude vs. Gemma 4 E2B | Overall (all concepts) | 125 | 86.4% (80.8–91.2) | 0.706 (0.597–0.811) |
| Gemma 4 31B vs. Gemma 4 E2B | Reasonableness | 25 | 68.0% (48.4–82.8) | 0.153 (-0.212–0.525) |
| Gemma 4 31B vs. Gemma 4 E2B | Cancer type/histology | 25 | 80.0% (60.9–91.1) | 0.545 (0.224–0.839) |
| Gemma 4 31B vs. Gemma 4 E2B | Prior treatment | 25 | 84.0% (65.3–93.6) | 0.655 (0.286–0.918) |
| Gemma 4 31B vs. Gemma 4 E2B | Biomarkers | 25 | 100.0% (86.7–100.0) | 1.000 |
| Gemma 4 31B vs. Gemma 4 E2B | Cancer burden | 25 | 92.0% (75.0–97.8) | -0.042 |
| Gemma 4 31B vs. Gemma 4 E2B | Overall (all concepts) | 125 | 84.8% (78.4–90.4) | 0.675 (0.545–0.795) |

*$\kappa = (p_{obs} - p_{chance}) / (1 - p_{chance})$. Where one answer category dominates a concept, $p_{chance}$ approaches 1 and κ becomes unstable even when raw agreement is high; this affects the cancer burden rows, in which 89–95% of annotations were “yes”. κ is undefined where both raters used a single category (0/0); exactly 0 where one rater used a single category, since chance agreement then equals observed agreement whatever its level (the cancer burden rows at 96% agreement); and slightly negative where observed agreement fell just below chance (92.0% vs 92.3%), indicating agreement no better than chance rather than systematic disagreement. κ = 1.000 in the biomarkers rows reflects perfect agreement on only two positive cells of 25. A κ interval is omitted where κ was undefined in more than 10% of bootstrap resamples.

# TRIPOD+LLM Checklist

Drafted by GPT 5.5 Pro Extended and GPT 5.6 Sol xhigh based on manuscript and supplement content.

| Section / Topic | Item Number | Checklist Item | Research Design | LLM Task | Reported on Page | Public notes |
| --- | --- | --- | --- | --- | --- | --- |
| **Title** | 2a | Identify the study as developing, fine-tuning, and/or evaluating the performance of an LLM, specifying the task, the target population, and the outcome to be predicted. | All | All | Main pp. 1-3 | The title identifies MatchMiner-AI as an open-source, privacy-preserving cancer clinical trial matching AI system. The abstract specifies that the platform uses open-weight LLMs for EHR summarization and trial target-population extraction, with custom retrieval/ranking models for cancer trial matching. |
| **Abstract** | 2b | Provide a brief explanation of the healthcare context, use case and rationale for developing or evaluating the performance of an LLM. | E,H | All | Main p. 2; pp. 4-6 | The abstract and describes oncology clinical trial matching, limited adult cancer trial participation, the manual burden on oncologists, privacy barriers to sharing PHI-trained AI models, and the rationale for an open-source synthetic-data approach. |
| **Objectives** | 2c | Specify the study objectives, including whether the study describes LLM development, tuning, and/or evaluation. | All | All | Main pp. 2, 6 | The objective is stated as developing an open-source AI platform for cancer clinical trial matching using fully synthetic EHR data, and evaluating it using real-world patient data. The abstract and introduction indicate model distillation/fine-tuning and evaluation. |
| **Methods** | 2d | Describe the key elements of the study setting. | All | All | Main pp. 2, 12, 16 | The study setting is Dana-Farber Cancer Institute, with DFCI EHR data accessed through the Oncology Data Retrieval System, DFCI clinical trial enrollment and standard-of-care treatment cohorts, and a production EHR-linked web interface for treating oncologists. |
| **Methods** | 2e | Detail all data used in the study, specify data splits and any selective use of data. | M,D,E | All | Main pp. 7-8, 12-15, 17-20, 29-30; Supp. pp. 2-22, 24-31 | The manuscript reports ClinicalTrials.gov trial data, extracted trial spaces, synthetic EHR histories/documents, DFCI trial-enrolled and standard-of-care evaluation cohorts, training/evaluation trial splits, a 10% standard-of-care patient-level sample, LLM-as-judge samples, and manual oncologist review samples. |
| **Methods** | 2f | Specify the name and version of LLM used. | All | All | Main pp. 7, 10-11, 14, 20, 32; Supp. pp. 28-31 | The manuscript names Gemma-4-31B-IT as the teacher LLM, Qwen3-0.6B-Embedding and ModernBERT-large for retrieval/reranking/classification components, and Claude Opus 4.8 as judge. |
| **Methods** | 2g | Briefly summarize the LLM-building steps, including any fine-tuning, reward modeling, reinforcement learning with human feedback (RLHF), etc. | M,D | All | Main pp. 2, 6-12; Supp. p. 23 | The abstract/methods summarize clinical trial space extraction, synthetic EHR generation, patient summarization, embedding-model fine-tuning, and TrialChecker/BoilerplateChecker distillation from teacher outputs and reasoning traces. |
| **Methods** | 2h | Describe the specific tasks performed by the LLMs (e.g., medical QA, summarization, extraction), highlighting key inputs and outputs used in the final LLM. | All | All | Main pp. 7-11, 14-15; Supp. pp. 2-22 | The manuscript and supplement describe trial-space extraction from eligibility criteria, synthetic longitudinal event and document generation, patient-history summarization from unstructured EHR text, patient-trial match-quality scoring, boilerplate exclusion assessment, LLM-as-judge evaluation, and cancer-type ascertainment for visualization. |
| **Methods** | 2i | Specify the evaluation datasets/populations used, including the endpoint evaluated, and detail whether this information was held out during training/tuning where relevant, and what measure(s) were used to evaluate LLM performance. | All | All | Main pp. 12-15, 17-20; Supp. pp. 24-31 | Evaluation populations include retrospective DFCI trial-enrolled and standard-of-care cohorts, LLM-as-judge patient-summary and patient-trial samples, and manual oncologist review samples. The manuscript reports that evaluation trials were not included in training and describes precision@N, MAP@N, AUROC, AUPRC, calibration, regression fidelity metrics, exact/within-1 agreement, kappa, and Spearman correlation. |
| **Results** | 2j | Give an overall report and interpretation of the main results. | All | All | Main pp. 3, 17-21, 32; Supp. pp. 24-31 | The abstract and results report improved MAP@20 versus baseline, performance of TrialChecker/BoilerplateChecker, LLM-as-judge and oncologist review findings, and comparison with a rules-based genomic trial-matching approach. |
| **Discussion** | 2k | Explicitly state any broader implications or concerns that have arisen in light of these results. | All | All | Main pp. 21-23 | The discussion addresses privacy-preserving shareability, open-source reproducibility, use of synthetic EHR text for clinical AI pipelines, the need for clinician oversight, nondeterminism, English-language limitation, subjectivity of "reasonable consideration," and external generalizability concerns. |
| **Other** | 2l | Give the registration number and name of the registry or repository (if relevant). | H | All | Not reported / not applicable | The study is a model development and evaluation study; no clinical trial or study registry entry is described. Public repositories for code, models, synthetic data, and demonstration apps are reported separately under data/code availability. |

| Section / Topic | Item Number | Checklist Item | Research Design | LLM Task | Reported on Page | Public notes |
|---|---|---|---|---|---|---|
| **Background** | 3a | Explain the healthcare context / use case (e.g., administrative, diagnostic, therapeutic, clinical workflow) and rationale for developing or evaluating the LLM, including references to existing approaches and models. | All | All | Main pp. 4-6 | The introduction describes oncology trial matching as a clinical workflow problem, reviews prior structured-data and LLM-based approaches, explains limitations of pre-summarized patient inputs and closed/proprietary systems, and motivates a synthetic-data, open-source pipeline. |
| **Background** | 3b | Describe the target population and the intended use of the LLM in the context of the care pathway, including its intended users in current gold standard practices (e.g., healthcare professionals, patients, public, or administrators). | E,H | All | Main pp. 2, 5-6, 12, 16 | The target population is adults with cancer receiving treatment at DFCI, represented by trial-enrolled and standard-of-care cohorts. The intended users are treating oncologists who retrieve ranked trial candidates through a lightweight EHR-linked interface; current practice is manual trial identification by oncologists and staff. |
| **Objectives** | 4 | Specify the study objectives, including whether the study describes the initial development, fine-tuning, or validation of an LLM (or multiple stages). | All | All | Main p. 6 | The final paragraph of the Introduction and the Methods overview specify development of MatchMiner-AI, synthetic-data generation, fine-tuning/distillation of pipeline components, and retrospective/LLM-as-judge/oncologist evaluations. |
| **Data** | 5a | Describe the sources of data separately for the training, tuning, and/or evaluation datasets and the rationale for using these data (e.g., web corpora, clinical research/trial data, EHR data). | All | All | Main pp. 7-8, 12-15; Supp. pp. 2-22 | Training data sources include ClinicalTrials.gov cancer trial eligibility criteria and teacher-LLM-generated synthetic longitudinal EHR text. Evaluation uses DFCI EHR text, DFCI trial-enrollment records, standard-of-care treatment starts, LLM-as-judge samples, and oncologist annotations. The rationale is privacy-preserving shareability and real-data transfer evaluation. |
| **Data** | 5b | Describe the relevant data points and provide a quantitative and qualitative description of their distribution and other relevant descriptors of the dataset (e.g., source, languages, countries of origin). | All | All | Main pp. 7, 17-18, 22, 29-30; Supp. pp. 24-31 | The manuscript gives counts for active cancer trials, extracted spaces, training trials/spaces, synthetic histories/documents, DFCI trial-enrolled and standard-of-care cohorts, demographic characteristics, disease centers, evaluation denominators, and stratified performance tables. It states the data and prompts are English-language. |
| **Data** | 5c | Specifically state the date of the oldest and newest item of text used in the development process (training, fine-tuning, reward modeling) and in the evaluation datasets. | M,D,E,H | All | Main pp. 7, 12, 14, 29-30 | The manuscript reports the ClinicalTrials.gov query date (October 2025), trial-enrollment and standard-of-care evaluation windows (January 2016-April 2024, with standard-of-care starts reported through 2023), and the May 2026 LLM-as-judge sample. |
| **Data** | 5d | Describe any data pre-processing and quality checking, including whether this was similar across text corpora, institutions, and relevant sociodemographic groups. | All | All | Main pp. 7-10, 12-15; Supp. pp. 1-22, 26-27 | Reported preprocessing includes parsing trial-space outputs, excluding non-compliant synthetic outputs, text augmentation of drug names/abbreviations, adding second-primary histories to a subset, chunked longitudinal summarization, embedding/reranking workflows, bootstrap CIs, and demographic-stratified analyses. |
| **Data** | 5e | Describe how missing and imbalanced data were handled and provide reasons for omitting any data. | M,D,E | All | Main pp. 12, 17, 24, 29-30 | The manuscript reports exclusion of non-compliant synthetic outputs, a 10% patient-level sample of the larger standard-of-care dataset, missing demographic categories in Table 1, and unavailable sharing of PHI-containing retrospective datasets. |
| **Analytical Methods** | 6a | Report the LLM name, version, and last date of training or use during inference. | All | All | Main pp. 7, 10-11, 14, 20, 32; Supp. pp. 28-31 | The manuscript names the main teacher, judge, embedding, classifier, and distilled LLM components, and provides some timing for ClinicalTrials.gov extraction and evaluation samples. |
| **Analytical Methods** | 6b | Specify the type of LLM architecture, and LLM building steps, including any hyperparameter tuning (e.g., temperature, length limits, penalties), prompt engineering, and any inference settings (e.g., seed, temperature, max token length) as relevant. | M,D,E | All | Main pp. 8-11; Supp. pp. 2-22 | The manuscript describes embedding, cross-encoder/reranking/regression, classification, and small distilled language model components. It gives prompt details in the supplement, chunking/context-length handling, and sampling settings for repeated inference. |
| **Analytical Methods** | 6c | Report details of LLM development process from text input to outcome generation, such as training, fine-tuning procedures, and alignment strategy (e.g., reinforcement learning, direct preference optimization, etc.) and alignment goals (e.g., helpfulness, honesty, harmlessness, etc.). | M,D | All | Main pp. 7-11; Supp. p. 23 | The manuscript describes hierarchical synthetic-data generation, fine-tuning TrialSpace using multiple-negatives ranking and CoSENT losses, and distilling TrialChecker/BoilerplateChecker onto ModernBERT-large. |
| **Analytical Methods** | 6d | Specify the initial and post-processed output of the LLM (e.g., probabilities, classification, unstructured text). | All | All | Main pp. 7-11, 13-14; Supp. pp. 2-19 | Outputs include unstructured trial-space text, boilerplate exclusions, synthetic clinical histories and documents, structured patient summaries, 0-5 match-quality scores, binary boilerplate exclusion outputs, classifier probabilities, embedding similarities, and post-processed retrieval/ranking metrics. |

| Section / Topic | Item Number | Checklist Item | Research Design | LLM Task | Reported on Page | Public notes |
|---|---|---|---|---|---|---|
| **Analytical Methods** | 6e | Provide details and rationale for any classification and how the probabilities were determined and thresholds identified. | All | C,OF | Main pp. 9, 11, 13-14; Supp. pp. 16-19, 24-27 | The manuscript describes the 0-5 match-quality score, binarization of reasonable consideration at score >=1, TrialChecker filtering at normalized score >=0.2, boilerplate exclusion labels/probabilities, and AUROC/AUPRC/calibration analyses. |
| **Analytical Methods** | 6f | Include metrics that capture the quality of generative outputs, such as consistency, relevance, and accuracy, compared to gold standards. | All | QA,IR,DG, SS,MT | Main pp. 14-15, 19-20; Supp. pp. 19-22, 28-31 | Generative-output quality is assessed using Claude Opus 4.8 as judge for patient summaries, with oncologist adjudication, and using agreement metrics for match-quality and boilerplate judgments. Supplemental Table 7 reports clinically faithful summaries and defect types. |
| **Analytical Methods** | 6g | Report the outcome metrics' relevance to downstream task at deployment time and correlation of metric to human evaluation of the text for the intended use. | E,H | All | Main pp. 13-15, 19-23; Supp. pp. 28-31 | Precision@N and MAP@N are tied to retrieval of clinically reasonable trial candidates, while agreement/kappa/Spearman metrics quantify fidelity to teacher and judge/oncologist assessments. The Discussion addresses clinical interpretability, human oversight, and prospective utility evaluation. |
| **LLM Output** | 7a | Clearly define the outcome, how the LLM predictions were calculated (e.g., formula, code, object, API), and evaluation metrics. | E,H | All | Main pp. 9, 11, 13-16; Supp. pp. 16-19, 24-31 | The manuscript defines the 0-5 reasonable-consideration score, boilerplate exclusion judgment, TrialChecker regression/binary outputs, MAP@N, precision@N, AUROC/AUPRC, calibration, RMSE/MAE/R-squared, kappa, and Spearman correlation. Code repositories are provided. |
| **LLM Output** | 7b | If outcome assessment requires subjective interpretation, describe the qualifications of the assessors, any instructions provided, relevant information on demographics of the assessors, and inter-assessor agreement. | All | All | Main pp. 12, 14-15, 19-20; Supp. pp. 19-22, 28-31 | Subjective assessments include Claude Opus 4.8 judgments, oncologist adjudication of patient summaries, and manual oncologist annotation of patient-trial candidates. The prompt/instructions are included in the supplement, and agreement statistics are reported. |
| **LLM Output** | 7c | Specify how performance was compared to other LLMs, humans, and other benchmarks or standards. | All | All | Main pp. 13, 18-21, 32; Supp. pp. 24, 28-31 | Performance is compared with a larger general-domain embedding baseline, a rules-based MatchMiner-Genomics approach, teacher LLM labels, Claude Opus 4.8 judge outputs, and medical oncologist annotations/agreement. |
| **Annotation** | 8a | If annotation was done, report how text was labeled, including providing specific annotation guidelines with examples. | All | All | Main pp. 14-15; Supp. pp. 19-22 | The manuscript describes manual oncologist annotation of patient-trial candidate components and includes detailed LLM-as-judge instructions in Supplemental Table 2, including example verdict categories for summary review. |
| **Annotation** | 8b | If annotation was done, report how many annotators labeled the dataset(s), including the proportion of data in each dataset that were annotated by more than 1 annotator. | All | All | Main pp. 14-15, 19-20; Supp. pp. 28-31 | The manuscript reports five oncologists annotating 25 patient-trial pairs, \dually assigned for inter-rater agreement, and reports LLM-as-judge sample sizes for summaries and patient-trial checks. |
| **Annotation** | 8c | If annotation was done, provide information on the background and experience of the annotators, and the inter-annotator agreement. | All | All | Main pp. 12, 15, 20; Supp. pp. 29-31 | The manuscript states that practicing medical oncologists performed candidate-match annotation and reports kappa-based inter-rater reliability with Opus/oncologist comparisons. |
| **Prompting** | 9a | If research involved prompting LLMs, provide details on the processes used during prompt design, curation, and selection. | All | All | Main pp. 7-12; Supp. pp. 1-22 | Prompting is central to trial-space extraction, synthetic history/document generation, summarization, match scoring, boilerplate checks, LLM-as-judge review, and cancer-type ascertainment. Full prompt pseudocode/instructions are provided in the supplement, and pilot feedback-driven prompt modifications are summarized. |
| **Prompting** | 9b | If research involved prompting LLMs, report what data were used to develop the prompts. | All | All | Main pp. 7-12; Supp. pp. 1-22 | Prompts were developed around ClinicalTrials.gov eligibility criteria, synthetic trial-space histories, unstructured EHR-like documents, patient summaries, trial spaces, and pilot oncologist feedback. |
| **Summarization** | 10 | Describe any preprocessing of the data before summarization. | All | SS | Main pp. 8-9; Supp. pp. 11-16 | The manuscript describes summarizing each patient full longitudinal history using chronological overlapping chunks of approximately 50,000 tokens with 500-token overlap, maintaining a running structured summary through successive chunks. The supplement includes the summarization prompt and earlier/full-record instructions. |
| **Instruction Tuning / Alignment** | 11 | If instruction tuning/alignment strategies were used, what were the instructions and interface used for evaluation, and what were the characteristics of the populations doing evaluation? | M,D | All | Not reported / not applicable | Evaluation interfaces included prompts for trial-space extraction, summarization, match-quality scoring, boilerplate exclusion, and LLM-as-judge assessment, with real DFCI evaluation populations described. Additional instruction tuning/alignment training per se was not performed. |
| **Compute** | 12 | Report compute, or proxies thereof (e.g., time on what and how many machines, cost on what and how many machines, inference time, floating-point operations per second (FLOPs)), required to carry out methods. | M,D,E | All | Main pp. 15-16 | The manuscript reports deployment-time compute proxies: TrialSpace, TrialChecker, and BoilerplateChecker under one billion parameters, consumer CPU/GPU feasibility, precomputed embeddings returning in seconds, and raw-note summarization on the order of three minutes. |

| Section / Topic | Item Number | Checklist Item | Research Design | LLM Task | Reported on Page | Public notes |
|---|---|---|---|---|---|---|
| **Ethics Approval** | 13 | Name the institutional research board or ethics committee that approved the study and describe the participant-informed consent or the ethics committee waiver of informed consent. | All | All | Main p. 12 | The Methods report DF/HCC IRB approval and waivers of informed consent due to retrospective design, large data volume, minimal risk, and infeasibility of re-contacting patients. |
| **Open Science** | 14a | Give the source of funding and the role of the funders for the present study. | All | All | Main pp. 3, 16, 25 | Funding sources are listed in the abstract and acknowledgments, and the Methods state that funders had no role in study design, data collection, analysis, interpretation, or writing. |
| **Open Science** | 14b | Declare any conflicts of interest and financial disclosures for all authors. | All | All | Main p. 24 | The manuscript states that the authors declare no competing interests. |
| **Open Science** | 14c | Indicate where the study protocol can be accessed or state that a protocol was not prepared. | H | All | Not reported | A formal study protocol or statement that no protocol was prepared is not currently described in the accepted manuscript text. |
| **Open Science** | 14d | Provide registration information for the study, including register name and registration number, or state that the study was not registered. | H | All | Not reported | No registry name or registration number is reported, and no explicit "not registered" statement is present. |
| **Open Science** | 14e | Provide details of the availability of the study data. | All | All | Main pp. 16, 24 | The manuscript states that synthetic data, model weights, and demonstration apps are available on Hugging Face, while retrospective DFCI evaluation data contain PHI and cannot be shared. |
| **Open Science** | 14f | Provide details of the availability of the code to reproduce the study results. | All | All | Main pp. 16, 24 | The manuscript provides GitHub repositories for training and inference code and Hugging Face resources for models and synthetic data. |
| **Public Involvement** | 15 | Provide details of any patient and public involvement during the design, conduct, reporting, interpretation, or dissemination of the study or state no involvement. | H | All | Not reported | The manuscript describes clinical oncologist pilot feedback and co-development but does not describe patient or public involvement or explicitly state that none occurred. |
| **Participants** | 16a | When using patient/EHR data, describe the flow of text/EHR/patient data through the study, including the number of documents/questions/participants with and without the outcome/label and follow-up time. | E,H | All | Main pp. 12-15, 17-20, 29-30; Supp. pp. 24-31 | The manuscript describes EHR data extraction for trial-enrolled and standard-of-care cohorts, summarization, retrieval/ranking, TrialChecker and BoilerplateChecker evaluation, and reports patient/event counts and evaluation denominators. |
| **Participants** | 16b | When using patient/EHR data, report the characteristics overall and, for each data source or setting, and for development/evaluation splits, including the key dates, key predictors, and sample size. | E,H | All | Main pp. 12, 18, 29-30; Supp. pp. 26-27 | Table 1 reports evaluation-cohort characteristics by trial-enrolled and standard-of-care groups, including sex, age, year, race, ethnicity, and disease center. The Results report key trial/patient counts and evaluation splits. |
| **Participants** | 16c | For LLM evaluation, show a comparison of the distribution of important predictors between development and evaluation data. | E,H | All | Not reported | The manuscript reports evaluation cohort characteristics and demographic-stratified performance. It also reports counts and settings for synthetic development data and real DFCI evaluation cohorts. |
| **Participants** | 16d | When using patient/EHR data, specify the number of participants and outcome events in each analysis (e.g., for LLM development, hyperparameter tuning, LLM evaluation). | E,H | All | Main pp. 7-8, 12, 14-15, 17-20, 29-30; Supp. pp. 24-31 | The manuscript gives patient/event counts for DFCI cohorts, trial-space counts, synthetic histories/documents, LLM-as-judge samples, oncologist annotation samples, and candidate counts in supplement tables. |
| **Performance** | 17 | Report LLM performance according to pre-specified metrics (see item 7a) and/or human evaluation (see item 7d). | All | All | Main pp. 18-21, 32; Supp. pp. 24-31 | Performance is reported for retrieval/ranking, TrialChecker, BoilerplateChecker, LLM-as-judge summary/trial-checking quality, and oncologist review, including confidence intervals and stratified tables. |
| **LLM Updating** | 18 | If applicable, report the results from any LLM updating, including the updated LLM and subsequent performance. | All | All | Main pp. 12, 17-21; Supp. p. 1 | The study describes iterative pipeline modification after pilot oncologist feedback and reports performance of the updated pipeline/components. Supplemental Table 1 lists feedback and changes. |
| **Interpretation** | 19a | Give an overall interpretation of the main results, including issues of fairness in the context of the objectives and previous studies. | All | All | Main pp. 18, 21-23 | The discussion interprets MatchMiner-AI as an open, privacy-preserving, synthetic-data-trained trial-matching pipeline and summarizes implications for reproducible clinical AI. Results note demographic-stratified performance with no systematic disparity. |
| **Limitations** | 19b | Discuss any limitations of the study and their effects on any biases, statistical uncertainty, and generalizability. | All | All | Main pp. 22-23 | Limitations include clinician oversight requirements, non-device/non-CDS framing, nondeterminism, English-only data, and need for external/multisite generalizability evaluation. Confidence intervals are reported. |

| Section / Topic | Item Number | Checklist Item | Research Design | LLM Task | Reported on Page | Public notes |
|---|---|---|---|---|---|---|
| **Usability of the LLM in context** | 19c | Describe any known challenges in using data for the specified task and domain context with reference to representation, missingness, harmonization, and bias. | E,H | All | Main pp. 5-6, 12-16, 18, 22-23 | The manuscript discusses barriers of raw longitudinal EHR text, lack of pre-generated summaries, PHI restrictions, data-sharing constraints, English-only data, and demographic-stratified performance. |
| **Usability of the LLM in context** | 19d | Define the intended use for the implementation under evaluation, including the intended input, end-user, level of autonomy/human oversight. | E,H | All | Main pp. 5-6, 16, 22 | Intended use is clinician-facing retrieval/ranking of cancer clinical trial options from longitudinal EHR text and trial eligibility documents, surfaced in an EHR-linked interface for oncologists. The Discussion states that MatchMiner-AI does not replace clinicians. |
| **Usability of the LLM in context** | 19e | If applicable, describe how poor quality or unavailable input data should be assessed and handled when implementing the LLM, i.e., what is the usability of the LLM in the context of current clinical care. | E,H | All | Not reported | The manuscript describes running on available EHR text and notes that DFCI PHI data cannot be shared; the workflow uses precomputed summaries and clinician review. It also notes that the system is not autonomous medical decision support. |
| **Usability of the LLM in context** | 19f | If applicable, specify whether users will be required to interact in the handling of the input data or use of the LLM, and what level of expertise is required of users. | E,H | All | Main pp. 5-6, 16, 22 | The system is intended for treating oncologists; patient summaries are precomputed and a lightweight web interface is surfaced in the production EHR so clinicians can retrieve trial options on demand. |
| **Usability of the LLM in context** | 19g | Discuss any next steps for future research, with a specific view to applicability and generalizability of the LLM. | All | All | Main p. 23 | Future work is described as evaluation and impact assessment among clinicians, research staff, and trialists, with retraining possible as LLMs improve and external generalizability requiring further evaluation. |

*Abbreviations in Research Design and LLM Task columns are retained from the TRIPOD+LLM checklist template.*